\documentclass[11pt]{article}

\usepackage[final]{acl}
\usepackage{enumitem}
\usepackage{times}
\usepackage{algorithm}
\usepackage{algpseudocode}
\usepackage{latexsym}
\usepackage{multirow}
\usepackage[T1]{fontenc}
\usepackage{amssymb}
\usepackage{subcaption}
\usepackage{tabularx}
\usepackage[utf8]{inputenc}
\usepackage{amsmath}
\usepackage{microtype}
\usepackage{booktabs}
\usepackage{inconsolata}
\usepackage{hyperref}

\usepackage{graphicx}
\usepackage{framed}
\usepackage[breakable]{tcolorbox}

\title{CURP: Codebook-based Continuous User Representation for \\Personalized Generation with LLMs}

\author{
  Liang Wang\textsuperscript{1*},
  Xinyi Mou\textsuperscript{1*},
  Xiaoyou Liu\textsuperscript{1},
  Xuanjing Huang\textsuperscript{3}
  Zhongyu Wei\textsuperscript{1,2\textdagger} \\
  \textsuperscript{1}School of Data Science, Fudan University \\
  \textsuperscript{2}Shanghai Innovation Institute \\
  \textsuperscript{3}School of Computer Science, Fudan University \\
  \texttt{liangwang25,xiaoyouliu25@m.fudan.edu.cn, xymou20,xjhuang,zywei@fudan.edu.cn}
}

\begin{document}
\maketitle
\begin{abstract}
User modeling characterizes individuals through their preferences and behavioral patterns to enable personalized simulation with Large Language Models (LLMs) in contemporary approaches. However, existing methods, whether personalized prompt-based or parameter-based methods, face challenges in balancing personalization quality against computational and data efficiency. We propose a novel framework CURP, which employs a bidirectional user encoder, a discrete prototype codebook and a dual-aggregator to extract both multi-dimensional user traits and query-aware user characteristic. This design enables plug-and-play personalization with a small number of trainable parameters (about 28M parameters, about 0.36\% of the total model size). Through extensive experiments, we show that CURP achieves superior performance and generalization compared to strong baselines on the language model personalization (LaMP) benchmark, while offering better interpretability and robustness. 
 

\end{abstract}

\section{Introduction}
User simulation aims to capture individual preferences and behaviors, allowing systems to generate personalized content and interactions~\cite{mou2024individual,xi2025rise,tseng2024two}.
Traditional user modeling typically relies on feature engineering and representation learning to perform coarse-grained attribute classification, limiting the ability to capture more complex user behaviors and preferences~\cite{purificato2024user,he2023survey}.
In contrast, LLMs can generate personalized content that reflects individual preferences, with promising applications in dialogue systems, recommendation, and healthcare interventions~\cite{he2025simulation,li2024agent,bao2024piors}.

\begin{figure}[!t]
  \centering
    \setlength{\belowcaptionskip}{-0.5cm}  \includegraphics[width=0.5\textwidth]{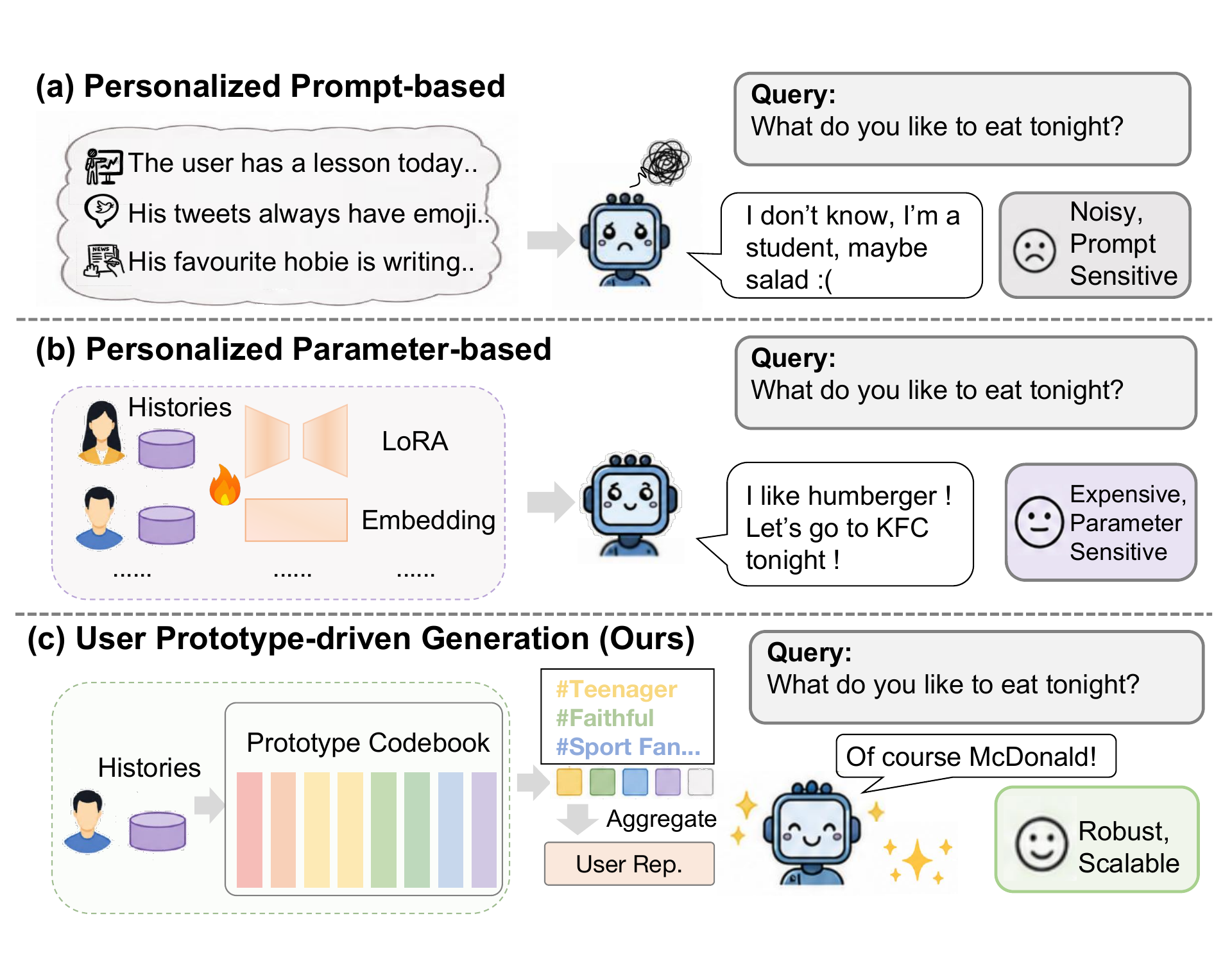} 
  \caption{Different paradigms for user response simulation.(a) \emph{Personalized Prompt-based}: prompting LLMs with text-based user behaviors, generalizable but noisy; (b) \emph{Personalized Parameter-based}: Finetuning parameters for each user, accurate but expensive; (c) \emph{Ours}: adopting prototype fusion based user representation via a codebook and an aggregator to guide user content generation, achieve a balance between two paradigms.}
  
  \label{fig:intro}
\end{figure}

Current approaches to LLM-based user simulation fall into two paradigms: \emph{personalized prompt} and \emph{Personalized Parameters}, as illustrated in Figure~\ref{fig:intro}(a) and (b) ~\cite{kumar2024longlamp,zhang2024guided,sun2024identity}.
Personalized prompt-based method injects histories and sometimes personas directly into the prompt to condition generation without modifying the model.
It is lightweight and general-purpose, but pure descriptions often contain noise and redundancy, leading to imprecise and stereotyped outputs.
In contrast, personalized parameter-based method adapts model weights or steers the latent representation using user-specific data, thereby internalizing personalized traits.
Although high fidelity is achieved, fine-tuning a separate model per user is computationally prohibitive and impractical at scale.
These limitations expose a fundamental tension between efficiency and personalization fidelity, raising a central question: \textbf{How can we achieve faithful user simulation while remaining computationally efficient and scalable?}

Self-Categorization Theory (SCT)~\cite{turner1987rediscovering} posits that identity is formed through a context-sensitive blend of shared social prototypes, like \emph{Teenager}, \emph{Faithful}, and \emph{Sport Fan}, rather than as a fixed and isolated self-concept. 
This theory provides a psychologically grounded and structured framework for user modeling, where identity is represented as meaningful and reusable prototypes. Motivated by SCT, we propose to model users in a shared latent prototype space, implemented as learnable prototypes ~\cite{van2017neural,deng2025onerec,arik2020protoattend}
, where each entry captures a salient behavioral or preference pattern. As shown in Figure~\ref{fig:intro}(c), a user might be encoded as a fusion of \emph{\#Teenager} + \emph{\#Faithful} + \emph{\#Sport Fan}. 
Notably, the shared space enables learning from aggregated signals across users, mitigating data sparsity while preserving individual differences through flexible prototype blending. This latent abstraction also improves robustness to noisy profile descriptions by mapping diverse user expressions into a common semantic space.

In this paper, we introduce \textbf{CURP} (\underline{C}odebook-based Continuous \underline{U}ser \underline{R}epresentation for \underline{P}ersonalized Generation with LLMs), a prototype-based framework for noise-robust and data-efficient personalized generation. At the core of CURP is a learnable \textbf{codebook} and a \textbf{dual-aggregator}.
CURP first encodes and maps user interaction histories into sparse combinations of shared prototypes with the codebook. The dual-aggregator then fuses these selected prototypes into complementary user representations, capturing both stable long-term preferences and dynamic query-specific intent.
The resulting personalized representation is subsequently projected and incorporated into the task prompt for personalized generation.
Based on this, training proceeds in two stages:
the \textbf{PCC} stage constructs codebooks from a large-scale behavioral pool, while the \textbf{UBA} stage learns the dual-aggregator and a lightweight adapter that maps the user representation into a frozen LLM’s space for personalized generation.
To validate our approach, we conduct experiments across six personalized tasks on LaMP benchmark~\cite{salemi2023lamp}. CURP consistently outperforms strong baselines, with ablation studies revealing key insights into its mechanisms. Our main contributions are summarized as follows: 

(1) We propose a novel framework that encodes each user as fused denoised prototypes and query-aware embeddings, replacing fragile textual descriptions and per-user parametric training.

(2) We introduce a two-stage training framework that enables the model to understand and utilize dual user representations without additional LLM training, making our approach model-agnostic.

(3) Extensive experiments across diverse personalization tasks demonstrate the effectiveness, scalability, and generalization ability of CURP, which achieves a practical balance between personalization fidelity and efficiency.

\begin{figure*}[t]
  \centering
  \includegraphics[width=0.98\textwidth]{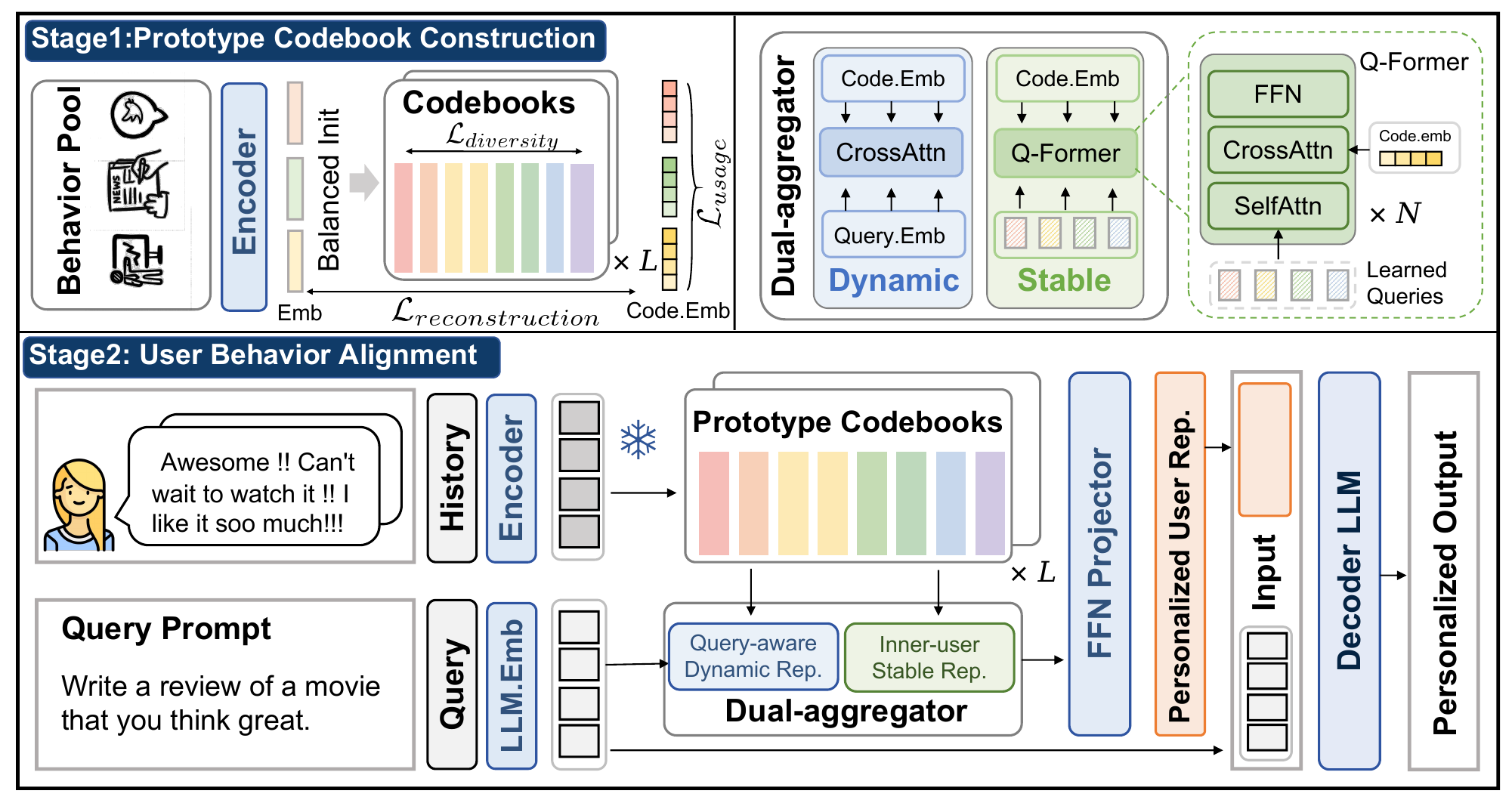}
  \caption{The overall framework of our proposed CURP. Our training procedure can be divided into two stages: Prototype Codebook Construction (PCC) and User Behaviour Aligning (UBA). Rep. is short for Representation.}
  \label{fig:frame}
\end{figure*}

\section{Related Works}
\subsection{User Modeling}

Prior to LLMs, user representations were typically constructed from explicit personas (e.g., questionnaires, self-disclosed attributes) or implicit behavioral signals, processed into vectors using TF-IDF~\cite{nanas2003building,elkahky2015multi,purificato2024user} for traditional machine learning or word embeddings~\cite{an2019neural,fazelnia2022variational} for deep learning. However, these approaches were largely limited to classification tasks and lacked generative capacity.

Recent advances in LLMs have enabled user simulation and personalized generation conditioned on user representations, either as natural language or learned embeddings~\cite{chen2024persona,li2024personalized,huber2025embeddingtoprefix}, moving beyond a generic one-size-fits-all paradigm toward more structured modeling of user characteristics~\cite{xu2025personalized,wang2023generative,wozniak2024personalized}.

\subsection{LLM-based Personalization}

Personalization methods for LLMs generally fall into two paradigms: \textit{prompt-based} approaches incorporate user profiles or interaction histories directly into the input context~\cite{richardson2023integrating,li2024llmasajudge,zhang2024guided}, but often struggle to interpret noisy signals at scale. \textit{Parameter-based} methods adapt model parameters via parameter-efficient fine-tuning~\cite{tan-etal-2024-democratizing,tan-etal-2024-personalized,zhang2024personalized} or personalized reward modeling~\cite{li2024personalized,bose2025lore}, albeit with higher computational costs. Recently, increasing attention has turned to \textit{latent space personalization}, where user features are encoded as soft prompts or adapters~\cite{doddapaneni2024user,jeongfactual,hebert2024persoma,huber2025embeddingtoprefix,zhang2025proper}. While some approaches employ discrete codebooks to represent semantic patterns in user utterances~\cite{he2025simulation,tang2024morpheus}, these codebooks focus primarily on dialog policy rather than capturing characteristics. PHF~\cite{wang2026PHF} further models user behavior hierarchically, both over time and across users, which fundamentally relies on high-quality and stable codebook-based practice representations. These studies highlight that developing interpretable, robust, and scalable user representations is both fundamental to effective personalization and remains a central open challenge.

\section{Methodology}
In this section, we first define the task and provide a detailed description of the proposed framework and training procedure. 
Given a set of users $\mathcal{U} = \{u_1, u_2, ..., u_n\}$, each user $u_i$ is defined as a tuple $u_i = \langle \mathcal{H}_i, \mathcal{Q}_i, \mathcal{R}_i \rangle$. Here, $\mathcal{H}_i$ represents the user's historical behaviors; $\mathcal{Q}_i$ is the current query for a scenario context; and $\mathcal{R}_i$ is the corresponding ground-truth from the user. Our goal is to leverage LLMs to accurately predict $\mathcal{R}_i$, given $\mathcal{H}_i$ and $\mathcal{Q}_i$. 


\subsection{Structure of CURP}
Figure~\ref{fig:frame} presents an overview of our proposed framework, which comprises four key components: the \textit{User Encoder} $\mathcal{E}$, the \textit{Prototype Codebook} $\mathcal{C}$, the \textit{Dual-Aggregator} $\mathcal{A}$ and the \textit{Decoder LLM} $\mathcal{D}$. 


\subsubsection{User Encoder}
To encode user historical behaviors, the fundamental source for personalization, we introduce a user encoder $\mathcal{E}$ that distills a user’s histories into a dense tensor representation. Unlike prior works that either couple the encoder with a task-specific adapter~\cite{hebert2024persoma,doddapaneni2024user} or require extensive task-specific data~\cite{ning2025user}, our encoder is built upon a pretrained encoder-only architecture with bidirectional attention, ensuring strong contextual modeling capacity while maintaining generalizability across diverse downstream tasks.
We encode historical behaviors using the encoder to obtain a unified representation of the user's history. This design addresses the limitations of naively concatenating raw texts, which introduces noise, expands context length, and fails to capture structured user characteristics.



\subsubsection{Codebook}
The codebook $\mathcal{C}$ is designed to achieve denoised structured behavior prototype for further aggregation. To construct this codebook, we first collect historical behaviors from a large-scale user pool, encoding them into dense tensors using the user encoder. These dense embeddings are then quantized through Product Quantization (PQ)~\cite{jegou2010product} into $L$ discrete indices from codebook vocabularies: each user event embedding is decomposed into $L$ subspaces, and each subspace is independently quantized to select an index from its own codebook. Through this process, each user event is abstracted as a combination of $L$ prototype essence. This design enables effective user modeling through the composition of learned prototypes, where each codebook entry captures a distinct aspect of user characteristics, eliminating the need for raw textual inputs, and provide informative but denoised representations for further aggregation.

\subsubsection{Dual-aggregator}


User preferences often exhibit stable long-term characteristics and dynamic context-dependent intent~\cite{mou2023uppam}. Modeling both aspects with a single representation may lead to information entanglement and reduced personalization quality. Therefore, we introduce a dual-aggregator $\mathcal{A}$ to decompose user modeling into two complementary branches: \textit{dynamic} and \textit{stable}. Let $\mathbf{BP}_i=\{\mathbf{bp}_{i,1}, \mathbf{bp}_{i,2}, \ldots, \mathbf{bp}_{i,LJ}\}$ denotes all behavior prototypes for user $u_i$. To capture user interests related to the current query, we employ a Multi-Head Cross-Attention mechanism~\cite{vaswani2017attention} that fuses the $\mathbf{BP}_i$ with the encoded task query $\mathcal{Q}_i$
, to acquire the query-aware dynamic user representation. Meanwhile, a stable user representation is constructed by compressing historical prototypes of varying lengths into a fixed-length vector, thereby preserving multi-dimensional cross-user characteristics. To achieve this, we utilize a multi-layer Q-Former~\cite{li2023blip} equipped with $M$ shared learnable queries. This dual-branch design effectively handles histories of arbitrary length while balancing the preservation of intrinsic user stability with query-specific task performance.


\subsubsection{Decoder}

To bridge the gap between the aggregated user representation space and the LLM decoder's embedding space, we employ a Multi-Layer Perceptron (MLP) to project the constructive embeddings into the decoder's embedding space.
With the task query $Q_i$ embedded via the decoder's input embedding layer, we combine the projected user embeddings with the query embeddings and feed the combined sequence into the decoder to perform personalized conditional generation, predicting the personalized response $R_i$. 
Our framework is decoder-agnostic, imposing no architectural constraints on the choice of decoder, which enables flexible application to various decoder architectures. The decoupled design that separates the codebook from the LLM also enables practical applications in cloud-edge collaboration and user privacy protection.

\subsection{Training CURP}
Based on the architecture described above, we implement a two-stage training pipeline: \textbf{Prototype Codebook Construction (PCC)} and \textbf{User Behavior Aligning (UBA)}.
\subsubsection{Stage 1: Prototype Codebook Construction} 
For all users $u_i \in \mathcal{U}$, we collect historical behaviors $\mathcal{H}_i$ from dataset $\mathcal{D}$ and encode them using the frozen encoder $\mathcal{E}$ to construct an embedding pool:
\begin{equation}
\mathcal{P} = \{\mathcal{E}(h_{i,j}) \mid h_{i,j} \in \mathcal{H}_i, \forall u_i \in \mathcal{U}\}.
\end{equation}
We partition each embedding $\mathbf{e} \in \mathbb{R}^d$ into $L$ subspaces: $\mathbf{e} = [\mathbf{e}^{(1)}, \ldots, \mathbf{e}^{(L)}]$, where $\mathbf{e}^{(l)} \in \mathbb{R}^{d/L}$. We initialize $L$ codebooks $\mathcal{C}=\{\mathcal{C}_1, \mathcal{C}_2,\dots, \mathcal{C}_L\}$ by applying balanced k-means algorithm~\cite{deng2025onerec,chen2025onesearch} for each subspaces to ensure uniform cluster sizes:



During quantization, each subspace embedding $\mathbf{e}^{(l)}$ is quantized by corresponding codebook:
\begin{equation}
\mathbf{e}_q = [\text{Quant}(\mathbf{e}^{(1)}, \mathcal{C}_1), \ldots, \text{Quant}(\mathbf{e}^{(L)}, \mathcal{C}_L)],
\end{equation}
where $\text{Quant}(\mathbf{e}^{(l)}, \mathcal{C}_l) = \arg\min_{\mathbf{c} \in \mathcal{C}_l} \|\mathbf{e}^{(l)} - \mathbf{c}\|^2$.
The training loss consists of three components:
\begin{equation}
\mathcal{L}_{\text{PCC}} = \alpha \mathcal{L}_{\text{reconstruction}} + \beta \mathcal{L}_{\text{diversity}} + \gamma \mathcal{L}_{\text{usage}},
\end{equation}
where $\mathcal{L}_{\text{reconstruction}}$ measures reconstruction error between original $\mathbf{e}$ and quantized $\mathbf{e}_q$, $\mathcal{L}_{\text{diversity}}$ and $\mathcal{L}_{\text{usage}}$ are used to encourage codebook diversity and usage rate to prevent codebook collapse.

More details for balanced K-Means and loss function can be found in Appendix~\ref{sec:kmeans} and ~\ref{sec:loss}.


\subsubsection{Stage 2: User Behavior Alignment} 

For each user $u_i$ with historical behaviors $\mathcal{H}_i = \{h_{i,1}, h_{i,2}, \ldots, h_{i,J}\}$, we independently encode and quantize each historical item $h_{i,j}$ as follows:
\begin{equation}
    \mathbf{e}_{i,j} = \mathcal{E}(h_{i,j}), \quad \mathbf{p}_{i,j} = PQ(\mathbf{e}_{i,j}, \mathcal{C}),
\end{equation}
where $\mathcal{E}(\cdot)$ denotes the encoder, and $PQ(\cdot, \mathcal{C})$ represents Product Quantization using the codebooks $\mathcal{C}$ learned in Stage 1. Here, $\mathbf{p}_{i,j}$ denotes the quantized prototype embedding for $h_{i,j}$. 

After that, we aggregate the prototypes into the merged dynamic and stable representations:
\begin{equation}
    \mathbf{R}_{dyn}= \mathbf{MHCA}(\mathbf{BP}_i, \mathcal{E}(\mathcal{Q}_i)),
\end{equation}
where $\mathbf{MHCA}$ 
denotes the Multi-Head Cross-Attention module.
\begin{equation}
    \mathbf{R}_{stb} = \textbf{Q-Former}(\mathbf{BP}_i, \mathbf{Q}_{learned}),
\end{equation}
where $\mathbf{Q}_{learned}$ represents $M$ learnable queries.

Finally, the dual representations are projected into the decoder's embedding space via MLPs ($f$) and prepended to the task query embedding. The final input sequence $\mathbf{x}_{\text{in}}$ is formulated as:
\begin{equation}
    \mathbf{x}_{\text{in}} = [f(\mathbf{R}_{dyn},\mathbf{R}_{stb}); \mathcal{D}_{\text{Emb}}(\mathcal{Q}_i)],
\end{equation}
where $\mathcal{D}_{\text{Emb}}$ denotes the embedding layer of the $\mathcal{D}$. The training objective maximizes the likelihood of generating the personalized response $Y_i$:
\begin{equation}
    \mathcal{L}_{\text{UBA}} = -\sum_{t=1}^{|Y_i|} \log P(y_t \mid \mathbf{x}_{\text{in}}, y_{<t}; \theta),
\end{equation}
where $\theta$ represents the trainable parameters, $y_t$ is the $t$-th token of the target response, and $|Y_i|$ is the total length of the response.

\subsection{Training details}

For the \textbf{PCC} stage, we randomly sample 100k user behaviors to balanced initialize the prototype codebooks $\mathcal{C}$. Each codebook consists 1,000 prototype entries for 4 subspaces by product quantization. 

For the \textbf{UBA} stage, we use the whole histories for each user. We employ Contriever~\cite{izacard2021unsupervised} as the base encoder $\mathcal{E}$ and Qwen-2.5-7B-Instruct~\cite{qwen2.5} as the decoder $\mathcal{D}$. We apply a 8-head MHCA for dynamic representation and a 2 layer Q-Former with 4 learnable queries for stable representation. A 2-layer MLP is used to project prototype embeddings into the LLM's embedding space. More details can be found in Appendix~\ref{sec:detailarg}.

\begin{table*}[]
\centering
\resizebox{\textwidth}{!}{%
\begin{tabular}{cccccccccccccccc}
\hline
\multirow{2}{*}{\textbf{Task}} & \multirow{2}{*}{\textbf{Metric}} & \multirow{2}{*}{\begin{tabular}{c}\textbf{Zero}\\\textbf{Shot}\end{tabular}} &    & \multicolumn{5}{c}{\textbf{Personalized prompt}} & \multicolumn{4}{c}{\textbf{Personalized parameters}}  & & \textbf{Ours} \\ \cline{4-8} \cline{10-13} \cline{15-15} 
 &  &  & ICL & BM25 & Dense & CoT & PAG &  & DiffMean & PROPER & SP & PPlug &  & CURP \\ \hline
\multicolumn{2}{c}{Lens \& Paras} & L & 9L & 9L & 9L & 9L & 2L &  & L\&4k* & L\&8M* & L\&4k & L\&40M &  & \textbf{L\&28M} \\ \hline
LaMP-1 & Acc$\uparrow$ & 0.514 & \underline{0.557} & \textbf{0.579} & 0.555 & 0.466 & 0.534 &  & - & - & 0.504 & 0.502 &  & 0.518 \\ \hline
\multirow{2}{*}{LaMP-2} & Acc$\uparrow$ & 0.345 & 0.445 & 0.431 & 0.448 & 0.423 & 0.384 &  & 0.345 & \underline{0.587} & 0.351 & 0.565 &  & \textbf{0.629} \\
 & F1$\uparrow$ & 0.191 & 0.291 & 0.281 & 0.300 & 0.080 & 0.110 &  & 0.182 & \underline{0.504} & 0.200 & 0.485 &  & \textbf{0.547} \\ \hline
\multirow{2}{*}{LaMP-3} & MAE$\downarrow$ & 0.502 & 0.368 & 0.314 & 0.284 & 0.381 & 0.577 &  & 0.580 & 0.284 & 0.524 & \textbf{0.258} &  & \underline{0.276} \\
 & RMSE$\downarrow$ & 0.839 & 0.714 & 0.651 & 0.604 & 0.725 & 0.940 &  & 0.923 & 0.597 & 0.867 & \textbf{0.567} &  & \underline{0.585} \\ \hline
\multirow{3}{*}{LaMP-4} & R-1$\uparrow$  & 0.138 & 0.160 & 0.168 & 0.177 & 0.143 & 0.136 &  & 0.136 & \underline{0.191} & 0.108 & 0.190 &  & \textbf{0.194} \\
 & R-L$\uparrow$   & 0.119 & 0.140 & 0.151 & 0.159 & 0.126 & 0.117 &  & 0.118 & \underline{0.171} & 0.092 & 0.168 &  & \textbf{0.173} \\
 & Sim$\uparrow$   & 0.845 & 0.852 & 0.854 & 0.855 & 0.849 & 0.842 &  & 0.844 & \underline{0.858} & 0.829 & 0.857 &  & \textbf{0.859} \\ \hline
\multirow{3}{*}{LaMP-5} & R-1$\uparrow$   & 0.411 & 0.432 & 0.440 & 0.435 & 0.448 & 0.376 &  & 0.396 & 0.469 & 0.292 & \underline{0.500} &  & \textbf{0.505} \\
 & R-L$\uparrow$   & 0.338 & 0.360 & 0.371 & 0.364 & 0.377 & 0.309 &  & 0.324 & 0.399 & 0.254 & \underline{0.444} &  & \textbf{0.449} \\
 & Sim$\uparrow$   & 0.876 & 0.884 & 0.885 & 0.884 & 0.887 & 0.870 &  & 0.874 & 0.890 & 0.864 & \underline{0.897} &  & \textbf{0.898} \\ \hline
\multirow{3}{*}{LaMP-7} & R-1$\uparrow$   & 0.457 & 0.428 & 0.446 & 0.447 & 0.364 & 0.416 &  & 0.463 & - & 0.451 & \underline{0.536} &  & \textbf{0.543} \\
 & R-L$\uparrow$   & 0.401 & 0.375 & 0.390 & 0.392 & 0.322 & 0.363 &  & 0.406 & - & 0.395 & \underline{0.477} &  & \textbf{0.483} \\
 & Sim$\uparrow$   & 0.903 & 0.894 & 0.898 & 0.898 & 0.881 & 0.889 &  & 0.900 & - & 0.899 & \underline{0.909} &  & \textbf{0.911} \\ \hline
\end{tabular}
}
\caption{Performance of all baselines and our model on six tasks. The best results are shown in \textbf{bold}, and the second-best are \underline{underlined}. The reported results represent the average of three independent runs.$\uparrow$ means higher is better. In Lens \& Paras, for personalized prompt methods we provide the relative average input length; for personalized parameters method we provide the number of personalized parameters as well. -indicates that the method is not adaptive to data formats. * denotes user-specific parameters.} 
\label{tab:main}
\end{table*}

\section{Experimental Setup}

\subsection{Datasets}

We conduct experiments on the LaMP benchmark~\cite{salemi2023lamp}. Following previous studies~\cite{tan-etal-2024-democratizing,tan-etal-2024-personalized}, we evaluate our method on 6
publicly available tasks, except for the Email Subject Generation (LaMP-6) for privacy. Generally, the six tasks consist of 3 classification tasks and 3 generation tasks: (1) LaMP-1: Citation Identification;
(2) LaMP-2: Movie Tagging; 
(3) LaMP-3: Product Rating; 
(4) LaMP-4: News Headline Generation; 
(5) LaMP-5: Scholarly Title Generation; 
(6) LaMP-7: Tweet Paraphrasing.

\subsection{Evaluation Metrics}
We use the default metrics align with LaMP benchmark to evaluate the performance of each tasks: accuracy (Acc) for LaMP-1, accuracy and F1 for LaMP-2, mean absolute error (MAE) and root mean squared error (RMSE) for LaMP-3, and ROUGE-1 (R-1) and ROUGE-L (R-L) for LaMP-4, LaMP-5, and LaMP-7. Besides, apart from the surface overlap ROUGE, we report the BERTScore (Sim) for semantic overlap for a better evaluation.

\subsection{Comparison Methods}
We compare CURP against a range of baselines, mainly within two broad categories: Personalized Prompt and Personalized Parameters.

\textbf{Personalized Prompt-based Methods :} These methods avoid parameter updates, providing efficiency at the cost of long context.
(1) \textbf{Zero Shot (ZS):} The LLM generates outputs solely based on the current input, serving as a non-personalized baseline.
(2) \textbf{In-Context Learning (ICL):} Historical user behaviors (default 8) as reference are appended to the prompt to personalize the LLM.
(3) \textbf{BM25:} BM25 algorithm to retrieve the most similar histories with the current query via surface overlap.
(4) \textbf{Dense:} Retrieve the most similar histories with query by dense Contriever embedding.
(5) \textbf{Chain-of-Thought (CoT)}~\cite{cot}: Prompts encourage step-by-step reasoning based on histories.
(6) \textbf{PAG}~\cite{zhang2024guided}: A persona profile is generated by Qwen-2.5-14B from the user's histories and prepended to the prompt.

\textbf{Personalized Parameter-based Methods:} These methods update parameters by supervision.

(1) \textbf{DiffMean}~\cite{zhang2025steer}: DiffMean represents each user as a vector in LLM's activation space for steered generation.
(2) \textbf{PROPER}~\cite{zhang2025proper}: PROPER further fine-tunes a personalized LoRA module for each user at varying granularities. While this approach maximizes the utilization of personal information, it incurs substantial computational costs
, which are known for their high resource demands.
(3) \textbf{Soft Prompt (SP)}~\cite{li2021sp}: Soft Prompt learns a shared trainable embedding prepended to the input sequence while keeping the model frozen.
(4) \textbf{PPlug}~\cite{liu2024llms+}: A soft prompt tuning based approach that aggregates the user's entire history into a single token via softmax weights derived from its similarity to the current query, functioning similarly to our dynamic representation.

\section{Experiment Result}

\subsection{Main Results}
The main results of inference on diverse scenarios are reported in Table \ref{tab:main}, and we can observe that:

(1) \textbf{Comparison of different methods}: Overall, CURP achieves the strongest overall performance by jointly modeling stable intrinsic user representations and query-aware dynamic adaptation. Personalized Prompt based methods like ICL, CoT and PAG yield suboptimal results, e.g. perform even worse than non-personalized ZS (R-1 0.457, R-L 0.401) in LaMP-7, primarily due to inflated context lengths and their sensitivity to noise in raw user histories, which also imply overthinking does not help personalization. While training-based methods improve performance, they incur substantial parameter costs for personalization. For example, although PROPER achieves competitive results in LaMP-2 and LaMP-4, it requires 8M parameters per user and the time efficiency is relatively much lower. DiffMean and SP perform way behind other baselines, and perform worse than ZS on LaMP-4 and LaMP-5, showing steering method and one general embedding hard to achieve personalization.
In contrast, our approach attains state-of-the-art results on more than 70\% of metrics across all tasks, with only about 30M parameters. This demonstrates a superior trade-off:\textbf{ CURP matches or exceeds the performance of heavier models while maintaining a compact and scalable architecture.}


(2) \textbf{Analysis across tasks}: The effectiveness of personalization methods varies with task characteristics, yet our approach delivers robust and leading performance consistently across diverse task types. On classification tasks, methods that better aggregate users' preferences regarding current query perform better. For instance, RAG-based methods BM25 and Dense show competitive performance even with training-based methods. In contrast, on generation tasks, the difference between methods is smaller than in classification tasks, while training methods PROPER, PPlug and CURP show a large advance than prompting-based methods. 
\textbf{Overall, our CURP model perform competitive performance in both task types.}



\subsection{Ablation Study}

\paragraph{Ablation on the Framework Component}

We first validate the necessity of each component through ablation studies by comparing performance after removing the corresponding module (Table~\ref{tab:Ablation}). Removing codebook quantization (\textit{w/o Quantize}) slightly improves performance but compromises robustness due to continuous embeddings (Section~\ref{sec:robust}). Eliminating either the dynamic or stable representation (\textit{w/o Dynamic/Stable}) causes clear degradation, confirming their complementary benefits. Replacing the Q-Former with mean pooling for stable representation (\textit{w/o Q-Former}) also yields suboptimal performance, showing that expressive feature fusion better captures user characteristics. Using separate MLPs (\textit{Dual MLP}) for different representations offers no noticeable gains while adding parameters. Therefore, we adopt a shared MLP as the projection module. Overall, each component plays an essential role and collectively contributes to the framework's effectiveness.

\begin{table}[t!]
\centering
\resizebox{\columnwidth}{!}{%
\begin{tabular}{ccccccccccccc}
\hline
\textbf{Task} &  & \multicolumn{2}{c}{\textbf{LaMP-2}} &  & \multicolumn{2}{c}{\textbf{LaMP-3}} &  & \multicolumn{2}{c}{\textbf{LaMP-4}} &  & \multicolumn{2}{c}{\textbf{LaMP-7}} \\ \cline{1-1} \cline{3-4} \cline{6-7} \cline{9-10} \cline{12-13} 
Metric &  & Acc$\uparrow$ & F1$\uparrow$ &  & MAE$\downarrow$ & RMSE$\downarrow$ &  & R-1$\uparrow$ & R-L$\uparrow$ &  & R-1$\uparrow$ & R-L$\uparrow$ \\ \hline
CURP &  & 0.629 & 0.547 &  & 0.276 & 0.585 &  & 0.194 & 0.173 &  & 0.543 & 0.483 \\ \hline
w/o Quantize &  & 0.634 & 0.548 &  & 0.275 & 0.601 &  & 0.198 & 0.177 &  & 0.544 & 0.483 \\ \hline
w/o Dynamic &  & 0.474 & 0.367 &  & 0.290 & 0.596 &  & 0.191 & 0.170 &  & 0.538 & 0.480 \\ \hline
w/o Stable &  & 0.539 & 0.436 &  & 0.287 & 0.607 &  & 0.188 & 0.168 &  & 0.528 & 0.469 \\ \cline{1-10} \cline{12-13} 
w/o Q-Former &  & 0.590 & 0.541 &  & 0.290 & 0.620 &  & 0.192 & 0.171 &  & 0.536 & 0.477 \\ \hline
Dual MLP &  & 0.604 & 0.558 &  & 0.282 & 0.605 &  & 0.194 & 0.175 &  & 0.545 & 0.485 \\ \hline
\end{tabular}
}
\caption{Overall performance of ablation models.}
\label{tab:Ablation}
\end{table}

\begin{figure}[t!]
    \centering
    \includegraphics[width=1.0\linewidth]{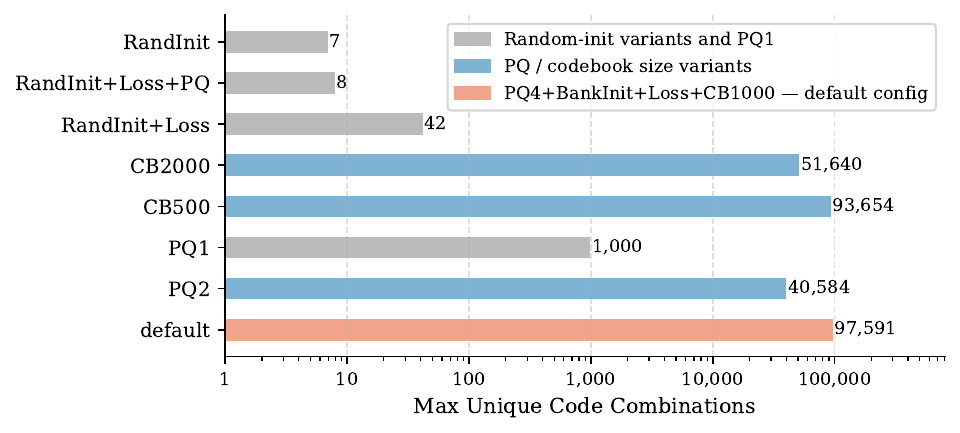}
    \caption{Maximum number of unique PQ code combination tuples observed across all training epochs for each codebook configuration on LaMP-2.}
    \label{fig:cb}
\end{figure}

\paragraph{Ablation on Codebook Design Choices} We further analyze how different design decisions affect codebook utilization and combination diversity. As illustrated in Figure~\ref{fig:cb}, we compare initialization strategies (e.g. Random), loss designs, vocabulary sizes (e.g. CB-500), and PQ subspace configurations (e.g. PQ1). Random initialization results in extremely low usage and can lead to codebook collapse, whereas balanced initialization with PQ4 and the proposed loss enables effective exploration of the codebook space. Overall, our final design (Balanced+PQ4+Loss) achieves both high entry usage, substantial combination diversity and cost-effective, outperforming alternatives significantly.

\begin{figure}[t!]
    \centering
    \includegraphics[width=1.0\linewidth]{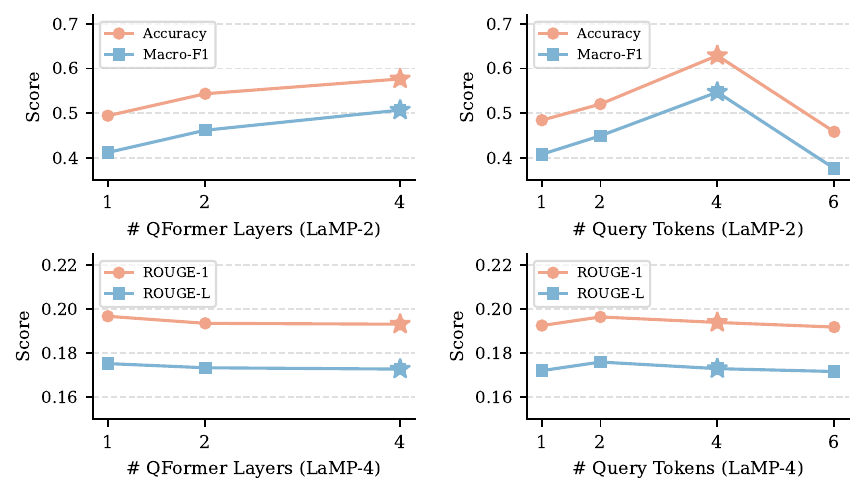}
    \caption{Ablation study on the number of Q-Former layers and learned queries on LaMP-2 and LaMP-4.}
    \label{fig:hyper}
\end{figure}

\paragraph{Ablation on Q-Former Layers and Learned Queries}
Finally, we investigate the impact of the number of Q-Former layers and learned queries on stable representation learning. As shown in Figure~\ref{fig:hyper}, 
for LaMP-2, the performance consistently improves as the number of Q-Former layers increases, whereas for LaMP-4, deeper Q-Former architectures lead to noticeable performance degradation. Regarding the number of learned queries, LaMP-2 achieves the best performance when the query number is set to 4, while the performance drops sharply when increasing it to 6. In contrast, LaMP-4 performs best with 2 learned queries, followed by a gradual decline as the query number increases.
Considering the trade-off across different tasks, we adopt a configuration with 2 Q-Former layers and 4 learned queries as the default setting.

\subsection{Further Analysis}
We further conduct additional analysis on LaMP-2 (classification) and LaMP-4 (generation).

\subsubsection{Robustness to Profile Paraphrase} \label{sec:robust}

\begin{figure}[t!]
    \centering
    \includegraphics[width=\linewidth]{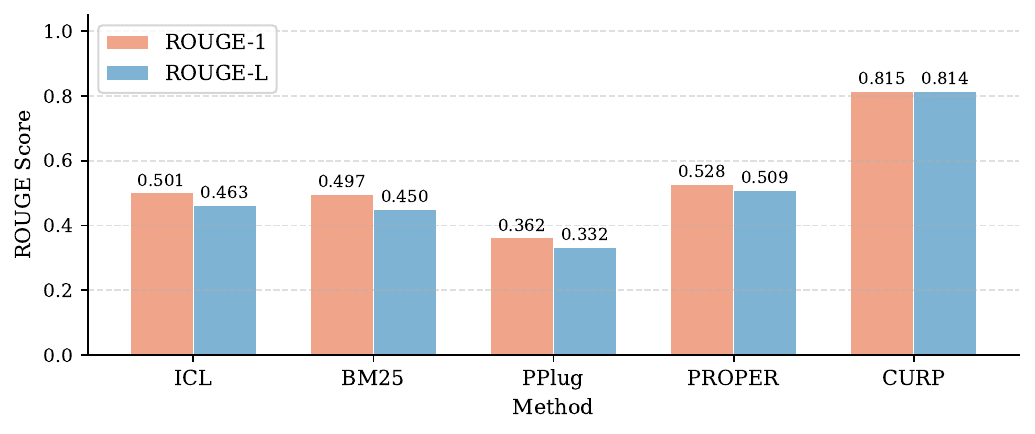}
    \caption{Robustness of personalized generation under profile paraphrase on LaMP-4.}
    \label{fig:paraphrase_rouge}
\end{figure}

In real-world applications, user profiles often come from diverse sources and may vary in wording or format, despite conveying the same underlying preferences. A robust personalization method should therefore rely on underlying semantic content for stable user understanding, rather than being distracted by surface-level lexical variations. To evaluate this, we design a paraphrase robustness test on LaMP-4 using 100 randomly sampled users. Specifically, we rewrite each user’s profile history using Qwen2.5-8B-Instruct while preserving its original meaning, and then run each method on both the original and paraphrased profiles. As shown in Figure~\ref{fig:paraphrase_rouge}, CURP achieves the highest consistency, demonstrating strong robustness to profile paraphrasing. This benefits from the quantization mechanism that encourages user representations to capture semantic-level patterns rather than superficial lexical features. In contrast, other methods achieve moderate consistency, indicating partial sensitivity to wording changes. Overall, these results demonstrate \textbf{CURP learns a more semantically grounded and robust user representation}.

\begin{figure}[t!]
    \centering
    \includegraphics[width=1\linewidth]{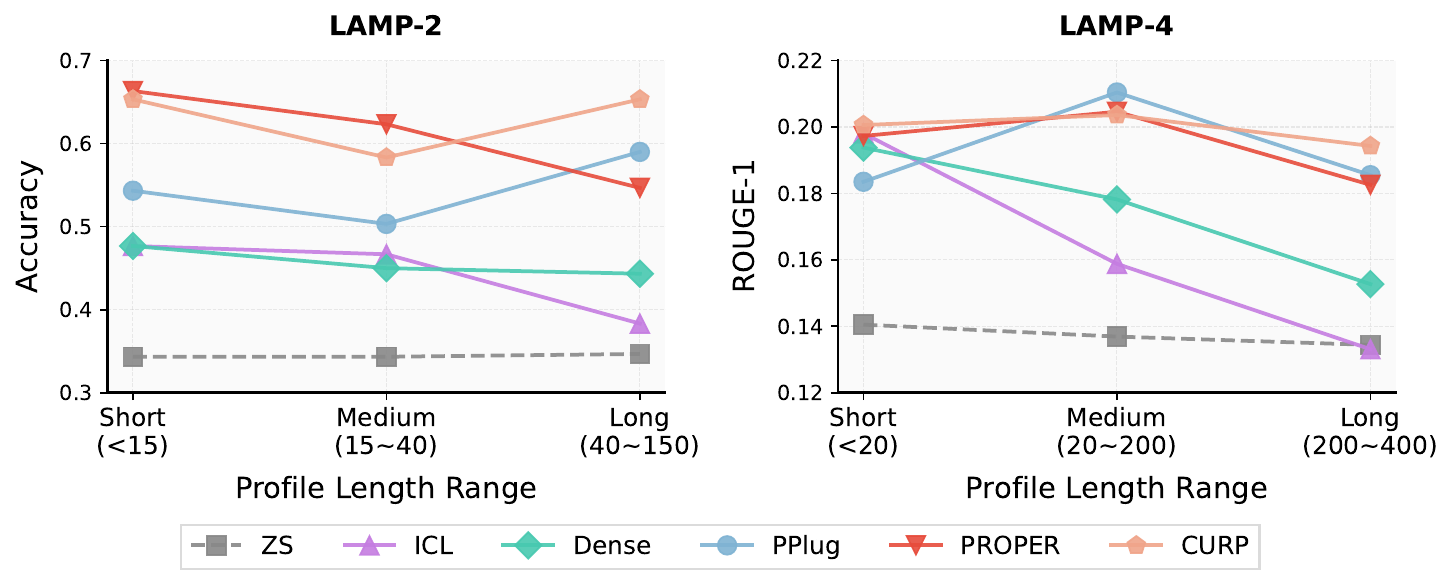}
    \caption{Performance of CURP with users with different history length.}
    \label{fig:lens}
\end{figure}

\subsubsection{Interpretability Analysis of Codebook}

To demonstrate the interpretability of our codebook-based representation, we analyze PQ indices from the LaMP-4 task. We observe a hierarchically structured representation: (1) \textbf{Topic-level consistency}: We first group samples by topic, e.g., Donald Trump, and examine their PQ indices. We observe that all samples within the same topic consistently share identical indices in subspaces indexed 40 and 101. This recurring pattern indicates that these subspaces function as a semantic anchor, capturing invariant, topic-level information. (2) \textbf{Within-topic variation}: Remaining subspaces capture nuanced semantic distinctions orthogonal to the topic. For instance, subspace 1 distinguishes content sources (e.g., legal proceedings vs. media coverage), while subspace 4 aligns with narrative framing (e.g., court challenges vs. personal impact). This indicates that our codebook effectively decomposes user profiles into stable topical cores and dynamic stylistic variations. Further visualizations and more details are provided in Appendix~\ref{sec:inter}.

\subsubsection{Generalization to Different Backbones}

\begin{figure}[t]
    \centering
    \includegraphics[width=\linewidth]{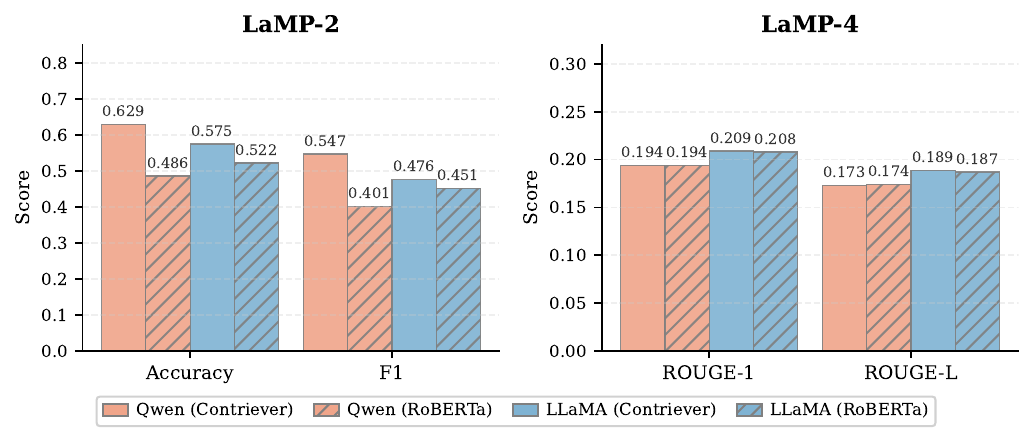}
    \caption{Generalization of CURP across different encoder and LLM backbones on LaMP-2 and LaMP-4.}
    \label{fig:general}
\end{figure}

To evaluate the generalization and modularity of CURP across different backbone models, we deploy the learned codebooks with three distinct $\mathcal{E}+\mathcal{D}$ combinations: \textit{RoBERTa+Qwen}, \textit{Contriever+LLaMA}, and \textit{RoBERTa+LLaMA}. When replacing the decoder, only the PBA stage needs to be re-executed, whereas changing the encoder requires rerunning both stages. As shown in Figure~\ref{fig:general}, CURP achieves consistently competitive performance across all combinations, indicating that \textbf{the learned codebook captures generalizable user characteristics that are not tied to any specific encoder or decoder architecture}. 

\subsubsection{Scalability Across Different User History Lengths}

To evaluate how personalization performance scales with increasing user history, we randomly sample 100 users with 3 different seeds from each of three profile-length groups (short, mid, and long), and guarantee similar difficulty by keeping ZS score stable, as shown in Table~\ref{fig:lens}. ICL and Dense perform weaker as user history length grows, as their context window are limited and can receive more noise. PROPER performs best on short profiles, likely benefiting from its LoRA-based task adaptation under limited user history. However, its advantage diminishes as profile length increases, likely because it weights all past interactions equally, failing to prioritize more relevant ones, and this may introduce noise and distort the direction. In contrast, PPlug struggles in the short-profile setting, due to its reliance on query-aware fusion, which becomes less effective when historical records are limited and misaligned. However, for CURP, \textbf{the codebook-based quantization compresses arbitrarily long user histories into compact and informative discrete representations while focusing on more relative histories in long history, allowing it to benefit consistently from additional history}. As a result, CURP achieves the best overall performance across both tasks.


\section{Conclusion}

In this work, we propose a novel CURP framework for personalized generation. To address the efficiency-fidelity trade-off in LLM-based personalized generation, we propose to model users as fused prototypes in both dynamic and stable representation. Experimental results demonstrate that our model achieves strong performance across multiple tasks, while further validating the generalization, interpretability, and robustness of our method.

\section{Limitations}

In this study we propose a novel prototype codebook-based approach for personalized LLM, however, several limitations remain that warrant future exploration. First, for LaMP benchmark personalization is basically rooted in behavior histories, however, missing the direct persona description of who the user is. Additionally, in real-world scenarios, users exhibit cross-domain behaviors that span multiple domains, which are currently unavailable in existing datasets. Second, completely discarding the textual content of historical behaviors inevitably leads to information loss regarding fine-grained behavioral patterns, such as habitual word usage.  Third, selecting appropriate evaluation metrics for personalized tasks is both challenging and crucial. Following previous works~\cite{salemi2023lamp,kumar2024longlamp,liu2024llms+,zhang2025steer} we report ROUGE-1 and ROUGE-L as well as BERTScore similarity to provide a comprehensive assessment, it remains difficult to deeply explore the degree of personalization. LLM as a judge is not involved due to their reliance on surface-level text for style assessment and not suitable for classification tasks. Since prompt-based methods already mimic historical writing and our framework avoids direct historical text input, using LLM as a judge would introduce information leakage, affecting evaluation objectivity. We look forward to future work that proposes more suitable metrics for evaluating LLM personalization.

\section{Ethical Considerations}

We address several ethical considerations in this work. First, all datasets used in our experiments are publicly available, and all models employed are open-source, ensuring that no user privacy information is involved in our research. We ensure that our experiments adhere to ethical guidelines and do not involve any ethical concerns.

\bibliography{custom}

@inproceedings{li2021sp,
  title={Prefix-tuning: Optimizing continuous prompts for generation},
  author={Li, Xiang Lisa and Liang, Percy},
  booktitle={Proceedings of the 59th Annual Meeting of the Association for Computational Linguistics and the 11th International Joint Conference on Natural Language Processing (Volume 1: Long Papers)},
  pages={4582--4597},
  year={2021}
}

@inproceedings{zhang2025proper,
  title={Proper: A progressive learning framework for personalized large language models with group-level adaptation},
  author={Zhang, Linhai and Wu, Jialong and Zhou, Deyu and He, Yulan},
  booktitle={Proceedings of the 63rd Annual Meeting of the Association for Computational Linguistics (Volume 1: Long Papers)},
  pages={16399--16411},
  year={2025}
}

@article{vaswani2017attention,
  title={Attention is all you need},
  author={Vaswani, Ashish and Shazeer, Noam and Parmar, Niki and Uszkoreit, Jakob and Jones, Llion and Gomez, Aidan N and Kaiser, {\L}ukasz and Polosukhin, Illia},
  journal={Advances in neural information processing systems},
  volume={30},
  year={2017}
}

@article{wang2026PHF,
  title={Beyond Isolated Behaviors: Hierarchical User Modeling for LLM
Personalization},
  author={Annoymous},
  journal={Under Review},
  year={2026}
}

@article{mou2024individual,
  title={From Individual to Society: A Survey on Social Simulation Driven by Large Language Model-based Agents},
  author={Mou, Xinyi and Ding, Xuanwen and He, Qi and Wang, Liang and Liang, Jingcong and Zhang, Xinnong and Sun, Libo and Lin, Jiayu and Zhou, Jie and Huang, Xuanjing and others},
  journal={arXiv preprint arXiv:2412.03563},
  year={2024}
}

@article{xi2025rise,
  title={The rise and potential of large language model based agents: A survey},
  author={Xi, Zhiheng and Chen, Wenxiang and Guo, Xin and He, Wei and Ding, Yiwen and Hong, Boyang and Zhang, Ming and Wang, Junzhe and Jin, Senjie and Zhou, Enyu and others},
  journal={Science China Information Sciences},
  volume={68},
  number={2},
  pages={121101},
  year={2025},
  publisher={Springer}
}

@article{tseng2024two,
  title={Two tales of persona in llms: A survey of role-playing and personalization},
  author={Tseng, Yu-Min and Huang, Yu-Chao and Hsiao, Teng-Yun and Chen, Wei-Lin and Huang, Chao-Wei and Meng, Yu and Chen, Yun-Nung},
  journal={arXiv preprint arXiv:2406.01171},
  year={2024}
}

@inproceedings{he2025simulation,
  title={Simulation-free hierarchical latent policy planning for proactive dialogues},
  author={He, Tao and Liao, Lizi and Cao, Yixin and Liu, Yuanxing and Sun, Yiheng and Chen, Zerui and Liu, Ming and Qin, Bing},
  booktitle={Proceedings of the AAAI Conference on Artificial Intelligence},
  year={2025}
}

@inproceedings{ning2025user,
  title={User-llm: Efficient llm contextualization with user embeddings},
  author={Ning, Lin and Liu, Luyang and Wu, Jiaxing and Wu, Neo and Berlowitz, Devora and Prakash, Sushant and Green, Bradley and O'Banion, Shawn and Xie, Jun},
  booktitle={Companion Proceedings of the ACM on Web Conference 2025},
  pages={1219--1223},
  year={2025}
}

@article{hebert2024persoma,
  title={PERSOMA: PERsonalized SOft ProMpt Adapter Architecture for Personalized Language Prompting},
  author={Hebert, Liam and Sayana, Krishna and Jash, Ambarish and Karatzoglou, Alexandros and Sodhi, Sukhdeep and Doddapaneni, Sumanth and Cai, Yanli and Kuzmin, Dima},
  journal={arXiv preprint arXiv:2408.00960},
  year={2024}
}

@article{li2024agent,
  title={Agent hospital: A simulacrum of hospital with evolvable medical agents},
  author={Li, Junkai and Lai, Yunghwei and Li, Weitao and Ren, Jingyi and Zhang, Meng and Kang, Xinhui and Wang, Siyu and Li, Peng and Zhang, Ya-Qin and Ma, Weizhi and others},
  journal={arXiv preprint arXiv:2405.02957},
  year={2024}
}

@article{bao2024piors,
  title={Piors: Personalized intelligent outpatient reception based on large language model with multi-agents medical scenario simulation},
  author={Bao, Zhijie and Liu, Qingyun and Guo, Ying and Ye, Zhengqiang and Shen, Jun and Xie, Shirong and Peng, Jiajie and Huang, Xuanjing and Wei, Zhongyu},
  journal={arXiv preprint arXiv:2411.13902},
  year={2024}
}

@article{bose2025lore,
  title={LoRe: Personalizing LLMs via Low-Rank Reward Modeling},
  author={Bose, Avinandan and Xiong, Zhihan and Chi, Yuejie and Du, Simon Shaolei and Xiao, Lin and Fazel, Maryam},
  journal={arXiv preprint arXiv:2504.14439},
  year={2025}
}

@article{li2024personalized,
  title={Personalized language modeling from personalized human feedback},
  author={Li, Xinyu and Zhou, Ruiyang and Lipton, Zachary C and Leqi, Liu},
  journal={arXiv preprint arXiv:2402.05133},
  year={2024}
}

@article{salemi2023lamp,
  title={Lamp: When large language models meet personalization},
  author={Salemi, Alireza and Mysore, Sheshera and Bendersky, Michael and Zamani, Hamed},
  journal={arXiv preprint arXiv:2304.11406},
  year={2023}
}

@article{chen2024persona,
  title={From persona to personalization: A survey on role-playing language agents},
  author={Chen, Jiangjie and Wang, Xintao and Xu, Rui and Yuan, Siyu and Zhang, Yikai and Shi, Wei and Xie, Jian and Li, Shuang and Yang, Ruihan and Zhu, Tinghui and others},
  journal={arXiv preprint arXiv:2404.18231},
  year={2024}
}

@article{xu2025personalized,
  title={Personalized generation in large model era: A survey},
  author={Xu, Yiyan and Zhang, Jinghao and Salemi, Alireza and Hu, Xinting and Wang, Wenjie and Feng, Fuli and Zamani, Hamed and He, Xiangnan and Chua, Tat-Seng},
  journal={arXiv preprint arXiv:2503.02614},
  year={2025}
}

@article{doddapaneni2024user,
  title={User embedding model for personalized language prompting},
  author={Doddapaneni, Sumanth and Sayana, Krishna and Jash, Ambarish and Sodhi, Sukhdeep and Kuzmin, Dima},
  journal={arXiv preprint arXiv:2401.04858},
  year={2024}
}

@inproceedings{li2023blip,
  title={Blip-2: Bootstrapping language-image pre-training with frozen image encoders and large language models},
  author={Li, Junnan and Li, Dongxu and Savarese, Silvio and Hoi, Steven},
  booktitle={International conference on machine learning},
  pages={19730--19742},
  year={2023},
  organization={PMLR}
}

@article{van2017neural,
  title={Neural discrete representation learning},
  author={Van Den Oord, Aaron and Vinyals, Oriol and others},
  journal={Advances in neural information processing systems},
  volume={30},
  year={2017}
}

@article{zhang2024personalized,
  title={Personalized llm response generation with parameterized memory injection},
  author={Zhang, Kai and Kim, Yejin and Liu, Xiaozhong},
  journal={arXiv preprint arXiv:2404.03565},
  year={2024}
}

@article{tang2024morpheus,
  title={Morpheus: Modeling role from personalized dialogue history by exploring and utilizing latent space},
  author={Tang, Yihong and Wang, Bo and Zhao, Dongming and Jin, Xiaojia and Zhang, Jijun and He, Ruifang and Hou, Yuexian},
  journal={arXiv preprint arXiv:2407.02345},
  year={2024}
}

@inproceedings{jeongfactual,
  title={Factual and Tailored Recommendation Endorsements using Language Models and Reinforcement Learning},
  author={Jeong, Jihwan and Chow, Yinlam and Tennenholtz, Guy and Hsu, ChihWei and Ghavamzadeh, Mohammad and Boutilier, Craig},
  booktitle={First Conference on Language Modeling},
  year={2024}
}

@inproceedings{tan-etal-2024-democratizing,
    title = "Democratizing Large Language Models via Personalized Parameter-Efficient Fine-tuning",
    author = "Tan, Zhaoxuan  and
      Zeng, Qingkai  and
      Tian, Yijun  and
      Liu, Zheyuan  and
      Yin, Bing  and
      Jiang, Meng",
    
    booktitle = "Proceedings of the 2024 Conference on Empirical Methods in Natural Language Processing",
    month = nov,
    year = "2024",
    address = "Miami, Florida, USA",
   
   
}

@article{sun2024identity,
  title={Identity-driven hierarchical role-playing agents},
  author={Sun, Libo and Wang, Siyuan and Huang, Xuanjing and Wei, Zhongyu},
  journal={arXiv preprint arXiv:2407.19412},
  year={2024}
}

@article{purificato2024user,
  title={User modeling and user profiling: A comprehensive survey},
  author={Purificato, Erasmo and Boratto, Ludovico and De Luca, Ernesto William},
  journal={arXiv preprint arXiv:2402.09660},
  year={2024}
}

@article{he2023survey,
  title={A survey on user behavior modeling in recommender systems},
  author={He, Zhicheng and Liu, Weiwen and Guo, Wei and Qin, Jiarui and Zhang, Yingxue and Hu, Yaochen and Tang, Ruiming},
  journal={arXiv preprint arXiv:2302.11087},
  year={2023}
}

@inproceedings{
huber2025embeddingtoprefix,
title={Embedding-to-Prefix: Continual Personalization with Large Language Models},
author={Bernd Huber and Ghazal Fazelnia and Andreas Damianou and Sebastian Peleato and Maksym Lefarov and Praveen Chandar and Marco De Nadai and Mounia Lalmas and Paul N. Bennett},
booktitle={AI That Keeps Up: NeurIPS 2025 Workshop on Continual and Compatible Foundation Model Updates},
year={2025}
}

@inproceedings{liu2024llms+,
    title = "{LLM}s + Persona-Plug = Personalized {LLM}s",
    author = "Liu, Jiongnan  and
      Zhu, Yutao  and
      Wang, Shuting  and
      Wei, Xiaochi  and
      Min, Erxue  and
      Lu, Yu  and
      Wang, Shuaiqiang  and
      Yin, Dawei  and
      Dou, Zhicheng",
    booktitle = "Proceedings of the 63rd Annual Meeting of the Association for Computational Linguistics (Volume 1: Long Papers)",
    month = jul,
    year = "2025",

}

@inproceedings{nanas2003building,
  title={Building and applying a concept hierarchy representation of a user profile},
  author={Nanas, Nikolaos and Uren, Victoria and De Roeck, Anne},
  booktitle={Proceedings of the 26th annual international ACM SIGIR conference on Research and development in informaion retrieval},
  pages={198--204},
  year={2003}
}

@inproceedings{elkahky2015multi,
  title={A multi-view deep learning approach for cross domain user modeling in recommendation systems},
  author={Elkahky, Ali Mamdouh and Song, Yang and He, Xiaodong},
  booktitle={Proceedings of the 24th international conference on world wide web},
  pages={278--288},
  year={2015}
}

@inproceedings{an2019neural,
  title={Neural news recommendation with long-and short-term user representations},
  author={An, Mingxiao and Wu, Fangzhao and Wu, Chuhan and Zhang, Kun and Liu, Zheng and Xie, Xing},
  booktitle={Proceedings of the 57th annual meeting of the association for computational linguistics},
  pages={336--345},
  year={2019}
}

@inproceedings{fazelnia2022variational,
  title={Variational user modeling with slow and fast features},
  author={Fazelnia, Ghazal and Simon, Eric and Anderson, Ian and Carterette, Benjamin and Lalmas, Mounia},
  booktitle={Proceedings of the Fifteenth ACM International Conference on Web Search and Data Mining},
  pages={271--279},
  year={2022}
}

@article{wang2023generative,
  title={Generative recommendation: Towards next-generation recommender paradigm},
  author={Wang, Wenjie and Lin, Xinyu and Feng, Fuli and He, Xiangnan and Chua, Tat-Seng},
  journal={arXiv preprint arXiv:2304.03516},
  year={2023}
}

@inproceedings{wozniak2024personalized,
  title={Personalized large language models},
  author={Wo{\'z}niak, Stanis{\l}aw and Koptyra, Bart{\l}omiej and Janz, Arkadiusz and Kazienko, Przemys{\l}aw and Koco{\'n}, Jan},
  booktitle={2024 IEEE International Conference on Data Mining Workshops (ICDMW)},
  pages={511--520},
  year={2024},
  organization={IEEE}
}

@article{kumar2024longlamp,
  title={Longlamp: A benchmark for personalized long-form text generation},
  author={Kumar, Ishita and Viswanathan, Snigdha and Yerra, Sushrita and Salemi, Alireza and Rossi, Ryan A and Dernoncourt, Franck and Deilamsalehy, Hanieh and Chen, Xiang and Zhang, Ruiyi and Agarwal, Shubham and others},
  journal={arXiv preprint arXiv:2407.11016},
  year={2024}
}

@inproceedings{tan-etal-2024-personalized,
    title = "Personalized Pieces: Efficient Personalized Large Language Models through Collaborative Efforts",
    author = "Tan, Zhaoxuan  and
      Liu, Zheyuan  and
      Jiang, Meng",
    booktitle = "Proceedings of the 2024 Conference on Empirical Methods in Natural Language Processing",
    month = nov,
    year = "2024",
}

@article{bm25,
author = {Robertson, Stephen and Zaragoza, Hugo},
title = {The Probabilistic Relevance Framework: BM25 and Beyond},
year = {2009},
issue_date = {April 2009},
publisher = {Now Publishers Inc.},
address = {Hanover, MA, USA},
volume = {3},
number = {4},
issn = {1554-0669},
url = {https://doi.org/10.1561/1500000019},
doi = {10.1561/1500000019},
journal = {Found. Trends Inf. Retr.},
month = apr,
pages = {333–389},
numpages = {57}
}

@article{zhang2024guided,
  title={Guided profile generation improves personalization with llms},
  author={Zhang, Jiarui},
  journal={arXiv preprint arXiv:2409.13093},
  year={2024}
}

@inproceedings{cot,
author = {Wei, Jason and Wang, Xuezhi and Schuurmans, Dale and Bosma, Maarten and Ichter, Brian and Xia, Fei and Chi, Ed H. and Le, Quoc V. and Zhou, Denny},
title = {Chain-of-thought prompting elicits reasoning in large language models},
year = {2022},
booktitle = {Proceedings of the 36th International Conference on Neural Information Processing Systems},

}

@misc{qwen2.5,
    title = {Qwen2.5: A Party of Foundation Models},
    author = {Qwen Team},
    month = {September},
    year = {2024}
}

@article{li2024llmasajudge,
      title   = {From Generation to Judgment: Opportunities and Challenges of LLM-as-a-judge},
      author  = {Dawei Li and Bohan Jiang and Liangjie Huang and Alimohammad Beigi and Chengshuai Zhao and Zhen Tan and Amrita Bhattacharjee and Yuxuan Jiang and Canyu Chen and Tianhao Wu and Kai Shu and Lu Cheng and Huan Liu},
      year    = {2024},
      journal = {arXiv preprint arXiv: 2411.16594}
    }

@article{richardson2023integrating,
  title={Integrating summarization and retrieval for enhanced personalization via large language models},
  author={Richardson, Chris and Zhang, Yao and Gillespie, Kellen and Kar, Sudipta and Singh, Arshdeep and Raeesy, Zeynab and Khan, Omar Zia and Sethy, Abhinav},
  journal={arXiv preprint arXiv:2310.20081},
  year={2023}
}

@article{zhang2025steer,
  title={Personalized Text Generation with Contrastive Activation Steering},
  author={Zhang, Jinghao and Liu, Yuting and Wang, Wenjie and Liu, Qiang and Wu, Shu and Wang, Liang and Chua, Tat-Seng},
  journal={arXiv preprint arXiv:2503.05213},
  year={2025}
}

@article{jegou2010product,
  title={Product quantization for nearest neighbor search},
  author={Jegou, Herve and Douze, Matthijs and Schmid, Cordelia},
  journal={IEEE transactions on pattern analysis and machine intelligence},
  volume={33},
  number={1},
  pages={117--128},
  year={2010},
  publisher={IEEE}
}

@article{izacard2021unsupervised,
  title={Unsupervised dense information retrieval with contrastive learning},
  author={Izacard, Gautier and Caron, Mathilde and Hosseini, Lucas and Riedel, Sebastian and Bojanowski, Piotr and Joulin, Armand and Grave, Edouard},
  journal={arXiv preprint arXiv:2112.09118},
  year={2021}
}

@article{chen2025onesearch,
  title={Onesearch: A preliminary exploration of the unified end-to-end generative framework for e-commerce search},
  author={Chen, Ben and Guo, Xian and Wang, Siyuan and Liang, Zihan and Lv, Yue and Ma, Yufei and Xiao, Xinlong and Xue, Bowen and Zhang, Xuxin and Yang, Ying and others},
  journal={arXiv preprint arXiv:2509.03236},
  year={2025}
}

@article{arik2020protoattend,
  title={Protoattend: Attention-based prototypical learning},
  author={Arik, Sercan O and Pfister, Tomas},
  journal={Journal of Machine Learning Research},
  volume={21},
  number={210},
  pages={1--35},
  year={2020}
}

@article{deng2025onerec,
  title={Onerec: Unifying retrieve and rank with generative recommender and iterative preference alignment},
  author={Deng, Jiaxin and Wang, Shiyao and Cai, Kuo and Ren, Lejian and Hu, Qigen and Ding, Weifeng and Luo, Qiang and Zhou, Guorui},
  journal={arXiv preprint arXiv:2502.18965},
  year={2025}
}

@book{turner1987rediscovering,
  title={Rediscovering the social group: A self-categorization theory.},
  author={Turner, John C and Hogg, Michael A and Oakes, Penelope J and Reicher, Stephen D and Wetherell, Margaret S},
  year={1987},
  publisher={basil Blackwell}
}

@inproceedings{mou2023uppam,
  title={Uppam: A unified pre-training architecture for political actor modeling based on language},
  author={Mou, Xinyi and Wei, Zhongyu and Zhang, Qi and Huang, Xuan-Jing},
  booktitle={Proceedings of the 61st Annual Meeting of the Association for Computational Linguistics (Volume 1: Long Papers)},
  pages={11996--12012},
  year={2023}
}

\appendix

\section{Training Procedure Details}

\subsection{Detailed Balanced K-Means Algorithm}\label{sec:kmeans}

To ensure that the codebook entries are utilized uniformly and represent distinct semantic clusters, we employ a \textit{Balanced K-Means} strategy for initialization. Unlike standard K-Means, which may suffer from codebook collapse (where few centroids attract most data points), our approach enforces a cardinality constraint on each cluster.

Formally, let $\mathcal{E}^{(l)} = \{\mathbf{e}_1^{(l)}, \dots, \mathbf{e}_N^{(l)}\}$ be the set of subspace embeddings for the $l$-th subspace. We aim to partition these embeddings into $K$ clusters $\{\mathcal{P}_k^l\}_{k=1}^K$ by minimizing the within-cluster sum of squares:

\begin{equation}
\min_{\{\mathcal{P}^l_k\}_{k=1}^K, \{\boldsymbol{\mu}_k^l\}_{k=1}^K} \sum_{k=1}^K \sum_{n \in \mathcal{P}^l_k} \| \mathbf{e}_n^{(l)} - \boldsymbol{\mu}_k^l \|^2
\end{equation}

subject to the balance constraint:
\begin{equation}
|\mathcal{P}^l_k| = \left\lfloor \frac{N}{K} \right\rfloor \quad \text{or} \quad \left\lceil \frac{N}{K} \right\rceil, \quad \forall k \in \{1, \ldots, K\}.
\end{equation}

Here, $\boldsymbol{\mu}_k^l$ denotes the centroid of the $k$-th cluster in the $l$-th codebook. The balance constraint ensures that each codebook entry is responsible for an approximately equal number of training samples, thereby maximizing the information capacity of the discrete codebook and preventing "dead" codes. We solve this constrained optimization problem using a swap-based iterative algorithm, as detailed in Algorithm~\ref{alg:balanced_kmeans}. For initialization, we run 100 iterations as default.

\begin{algorithm*}[t]
\caption{Balanced K-Means for Codebook Update}
\label{alg:balanced_kmeans}
\begin{algorithmic}[1]
\Require Subspace embeddings $\mathcal{E}^{(l)} = \{\mathbf{e}_n^{(l)}\}_{n=1}^N$, Codebook centroids $\mathcal{C}^l = \{\boldsymbol{\mu}_k^l\}_{k=1}^K$, Max iterations $T_{max}$
\Ensure Updated centroids $\mathcal{C}^l$, Assignments $\{\mathcal{P}_k^l\}_{k=1}^K$

\State Initialize assignments $\{\mathcal{P}_k^l\}_{k=1}^K$ randomly satisfying the balance constraint $|\mathcal{P}_k^l| \approx N/K$
\For{$t = 1$ to $T_{max}$}
    \State \textit{// Step 1: Update Centroids}
    \For{$k = 1$ to $K$}
        \State $\boldsymbol{\mu}_k^l \leftarrow \frac{1}{|\mathcal{P}_k^l|} \sum_{n \in \mathcal{P}_k^l} \mathbf{e}_n^{(l)}$
    \EndFor
    
    \State \textit{// Step 2: Compute Distance Matrix}
    \State Compute $D \in \mathbb{R}^{N \times K}$ where $D_{nk} = \|\mathbf{e}_n^{(l)} - \boldsymbol{\mu}_k^l\|^2$
    
    \State \textit{// Step 3: Balanced Re-assignment (Swap-based)}
    \State Initialize priority queue $Q$ with samples sorted by assignment cost
    \State $changes \leftarrow \text{true}$
    \While{$Q \neq \emptyset$ and $changes$}
        \State Select sample $n$ from $Q$ assigned to cluster $a$
        \State Find best alternative cluster $b \neq a$ minimizing distortion while maintaining balance
        \If{swapping $n$ from $a$ to $b$ reduces total distortion}
            \State Move $n$ from $\mathcal{P}_a^l$ to $\mathcal{P}_b^l$
            \State Update $\boldsymbol{\mu}_a^l$ and $\boldsymbol{\mu}_b^l$ incrementally
            \State $changes \leftarrow \text{true}$
        \Else
            \State $changes \leftarrow \text{false}$
        \EndIf
    \EndWhile
    
    \If{convergence criterion met}
        \State \textbf{break}
    \EndIf
\EndFor
\State \Return $\mathcal{C}^l, \{\mathcal{P}_k^l\}_{k=1}^K$
\end{algorithmic}
\end{algorithm*}

\subsection{Detailed Codebook Loss Composition}\label{sec:loss}

The training objective for the codebook module consists of three complementary components: reconstruction fidelity, codebook diversity, and usage uniformity. The total loss is defined as:
\begin{equation}
\mathcal{L}_{codebook} = \alpha\mathcal{L}_{\text{reconstruction}} + \beta \mathcal{L}_{\text{div}} + \gamma \mathcal{L}_{\text{usage}}
\end{equation}

\paragraph{Reconstruction Loss.}
To ensure that the quantized vectors faithfully approximate the original continuous embeddings, we minimize the squared Euclidean distance between the encoder output $\mathbf{e}_q$ and the selected codebook vector $\mathbf{e}$. Since the $\arg\min$ operation in vector quantization is non-differentiable, we employ the Straight-Through Estimator (STE) to allow gradient flow from the decoder to the encoder:
\begin{equation}
\mathcal{L}_{\text{reconstruction}} = \| \mathbf{e}_q - \text{sg}(\mathbf{e}) \|^2 + \| \text{sg}(\mathbf{e}_q) - \mathbf{e} \|^2
\end{equation}
where $\text{sg}(\cdot)$ is the stop-gradient operator. The second term acts as a commitment loss, encouraging the encoder outputs to stay close to the codebook embeddings they select.

\paragraph{Diversity Loss.}
To prevent multiple codebook entries from converging to the same semantic center (redundancy), we introduce a diversity loss that maximizes the minimum pairwise distance between centroids. Directly optimizing the hard minimum is unstable; thus, we use a smooth surrogate based on the hyperbolic tangent function:
\begin{equation}
\mathcal{L}_{\text{div}} = -\tau \cdot \text{tanh}\left(\frac{1}{\tau} \min_{i \neq j} \| \boldsymbol{\mu}_i - \boldsymbol{\mu}_j \| \right)
\end{equation}
where $\tau$ is a temperature parameter controlling the smoothness of the approximation. This term encourages the codebook to span a larger volume in the latent space, improving its ability to represent out-of-distribution or diverse user profiles.

\paragraph{Usage Loss.}
Codebook collapse, where only a small subset of entries is actively used, is a common failure mode in VQ-based models. We propose a composite usage loss $\mathcal{L}_{\text{usage}}$ that penalizes uneven utilization through three mechanisms:
\begin{equation}
\begin{split}
\mathcal{L}_{\text{usage}} = & \underbrace{\text{Var}(\{n_k\}_{k=1}^K)}_{\text{Variance Penalty}} \\
& + \underbrace{\left(1 - \frac{|\{k: n_k > 0\}|}{K}\right)}_{\text{Coverage Penalty}} \\
& + \underbrace{\left(1 - \frac{H(\{p_k\}_{k=1}^K)}{\log K}\right)}_{\text{Entropy Penalty}}
\end{split}
\end{equation}
Here, $n_k$ is the number of times the $k$-th codebook entry is selected in the current batch, and $p_k = n_k / \sum_j n_j$ is the empirical probability of usage. 
\begin{itemize}
    \item The \textit{Variance Penalty} minimizes the variance in usage counts, pushing towards equal frequency.
    \item The \textit{Coverage Penalty} directly maximizes the number of unique active codes (cardinality).
    \item The \textit{Entropy Penalty} maximizes the entropy of the usage distribution $H(\{p_k\})$, normalized by $\log K$ to range between $[0,1]$. A uniform distribution yields maximum entropy.
\end{itemize}
Together, these terms ensure robust and comprehensive utilization of the codebook capacity. The gradient is optimized after a large batch size of 1024 items with AdamW optimizer.

\section{Dataset Details}
 The format of question, answer and user
 histories of the tasks are shown in Table~\ref{tab:data}. We do data filter to make sure the length distribution. During the UBA stage, task-specific training data is utilized exclusively for the ID component, while other modules remain task-agnostic. 
Detailed statistics and example input output pair for all six tasks are provided in Table 7.

\begin{table*}[t!]
\centering
\caption{Dataset statistics and input--output formats for different tasks in the LaMP benchmark. “\#” denotes the number of instances, “Cls” indicates classification tasks, and placeholders in braces \{\} are replaced with dataset-specific values.}

\resizebox{\textwidth}{!}{%
\begin{tabular}{ccccccccc}
\hline
\textbf{Task} & \textbf{\#Train} & \textbf{\#Val} & \textbf{\#Len IN} & \textbf{\begin{tabular}[c]{@{}c@{}}\#Cls or\\ \#Len OUT\end{tabular}} & \textbf{Len HIST} & \textbf{Input Format}                                                                                                                                                                                     & \textbf{\begin{tabular}[c]{@{}c@{}}Output \\ Format\end{tabular}}                           & \textbf{\begin{tabular}[c]{@{}c@{}}History \\ Format\end{tabular}}             \\ \hline
LaMP-1        & 6,542            & 1,500          & 51.43           & 2                                                                    & 84.15             & \begin{tabular}[c]{@{}c@{}}For an author who has written the paper with title ``\{title\}'', which reference \\ is related? Just answer with <1> or <2>. <1>: ``\{ref1\}'' <2>: ``\{ref2\}''\end{tabular} & <2>                                                                                         & \begin{tabular}[c]{@{}c@{}}Title: \{title\} \\ Abstract: \{abs\}\end{tabular}  \\ \hline
LaMP-2        & 5,073            & 1,410          & 92.39           & 15                                                                   & 86.76             & \begin{tabular}[c]{@{}c@{}}Which tag does this movie relate to among the following tags? Just \\ answer with the tag name. tags: [sci-fi, action, ...] description: \{movie\}\end{tabular}                & romance                                                                                     & \begin{tabular}[c]{@{}c@{}}Description: \{des\}\\  Tag: \{tag\}\end{tabular}   \\ \hline
LaMP-3        & 20,000           & 2,500          & 128.18          & 5                                                                    & 185.40            & \begin{tabular}[c]{@{}c@{}}What is the score of the following review on a \\ scale of 1 to 5? Just answer with 1--5. review: \{review\}\end{tabular}                                                      & 5                                                                                           & \begin{tabular}[c]{@{}c@{}}Text: \{Review\} \\ Score: \{score\}\end{tabular}   \\ \hline
LaMP-4        & 12,500           & 1,500          & 29.97           & 10.07                                                                & 204.59            & Generate a headline for the following article: \{article\}                                                                                                                                                & How I Got 'Rich'                                                                            & \begin{tabular}[c]{@{}c@{}}title: \{title\} \\ text: \{article\}\end{tabular}  \\ \hline
LaMP-5        & 14,682           & 1,500          & 162.34          & 9.71                                                                 & 87.88             & Generate a title for the following abstract of a paper: \{abstract\}                                                                                                                                      & \begin{tabular}[c]{@{}c@{}}Distributed Partial \\ Clustering\end{tabular}                   & \begin{tabular}[c]{@{}c@{}}title: \{title\} \\ text: \{abstract\}\end{tabular} \\ \hline
LaMP-7        & 13,437           & 1,498          & 29.72           & 16.96                                                                & 15.71             & Paraphrase the following tweet without any explanation: \{tweet\}                                                                                                                                         & \begin{tabular}[c]{@{}c@{}}gotta make the most of \\ my last full day in ktown\end{tabular} & text: \{tweet\}                                                                \\ \hline
\end{tabular}
}
\label{tab:data}
\end{table*}

\begin{table}[]
\resizebox{\columnwidth}{!}{%
\begin{tabular}{lccc}
\hline
 & \multicolumn{3}{c}{LaMP-4} \\ \cline{2-4} 
\textbf{Method} & \textbf{Train Time} & \textbf{Infer Time} & \textbf{Memory} \\ \hline
Direct & --- & 3:36 & 14G \\
BM25 & --- & 6:16 & 18G \\
CoT & --- & 43:38 & 18G \\
PAG & --- & 12:20 & 16G \\
PPlug & 7:43 & 5:02 & 17G \\
DiffMean & --- & $\approx$640 min$^\dagger$ & 14G \\
PROPER & $\approx$57 min$^\dagger$ & $\approx$1056 min$^\dagger$ & 21G \\ \hline
\textbf{CURP (ours)} & \textbf{7:40} & \textbf{4:28} & 17G \\ \hline
\end{tabular}
}
\caption{Efficiency comparison between methods.}
\label{tab:eff}
\end{table}

\section{Efficiency}
We provide an efficiency experiment of the training and inference time on the same setting as shown in Table~\ref{tab:eff}. Our CURP achieve a high efficient, especially compare with DiffMean and PROPER.

\begin{table}[t]
\centering
\caption{Performance comparison of CURP with and without LoRA fine-tuning across LaMP benchmarks. Arrows indicate direction of improvement ($\uparrow$ higher is better, $\downarrow$ lower is better).}
\label{tab:lora_ablation}
\resizebox{\columnwidth}{!}{%
\begin{tabular}{lcccccccc}
\toprule
\multirow{2}{*}{\textbf{Model}} & \multicolumn{2}{c}{\textbf{LaMP-2}} & \multicolumn{2}{c}{\textbf{LaMP-3}} & \multicolumn{2}{c}{\textbf{LaMP-4}} & \multicolumn{2}{c}{\textbf{LaMP-7}} \\
\cmidrule(lr){2-3} \cmidrule(lr){4-5} \cmidrule(lr){6-7} \cmidrule(lr){8-9}
 & Acc$\uparrow$ & F-1$\uparrow$ & MSE$\downarrow$ & RMSE$\downarrow$ & R-1$\uparrow$ & R-L$\uparrow$ & R-1$\uparrow$ & R-L$\uparrow$ \\
\midrule
CURP      & 0.629 & 0.547 & 0.276 & 0.585 & 0.194 & 0.173 & \textbf{0.543} & \textbf{0.483} \\
w/ LoRA   & \textbf{0.633} & \textbf{0.561} & \textbf{0.270} & \textbf{0.574} & \textbf{0.214} & \textbf{0.192} & 0.538 & 0.481 \\
\bottomrule
\end{tabular}
}
\end{table}

\section{Hyperparameter Settings}\label{sec:detailarg}
All experiments are conducted on 8 NVIDIA 4090 GPUs with a batch size of 2 per GPU. We utilize DeepSpeed ZeRO Stage 2 to enable efficient memory usage and distributed training. The learning rate is set to 1e-4 with weight decay of 0.01 and max gradient norm of 1.0. We use gradient accumulation with 4 steps and train for 3 epochs. The MLP projection layer consists of two linear layers: 768-3584 with GELU activation, followed by 3584-3584. We use a 2 layer Q-Former with 4 learnable queries. The head for attention is set to 8. For the codebook training, we set the codebook weight to 1.0, diversity weight to 0.1, and usage weight to 0.3. We employ Contriever as the encoder and Qwen-2.5-7B-Instruct as the decoder. During inference, we set do\_sample as false to ensure replicability and the reliability of our result. The random seeds are 42, 0, 1 for multiple runs. We use the default val dataset for evaluation on LaMP benchmark. The default optimizer is AdamW for its stable performance. We use rouge\_score to compute ROUGE.

\section{Prompt Templates}

We provide detailed prompt templates used for encoding user history across different tasks.

In this section, we detail the prompt templates used for each LaMP benchmark. The prompts are constructed using three components: 
\begin{itemize}
    \item \textbf{Static Context}: Encodes long-term user preferences/profile (\texttt{stable\_str}).
    \item \textbf{Dynamic Context}: Encodes retrieved relevant history (\texttt{dynamic\_token}).
    \item \textbf{Instruction}: The specific task description and output constraint.
\end{itemize}

Note that \texttt{\{input\_text\}} represents the specific input for the current instance (e.g., the paper abstract, the movie description, or the article body). Ablation studies may remove either the Static or Dynamic context lines.

\subsection{LaMP-1: Scholar Profile Matching}
\begin{tcolorbox}[colback=gray!5, colframe=gray!40, title=Prompt Template, fonttitle=\bfseries]
\begin{verbatim}
The research expertise and dominant 
scientific topics that define this 
scholar's work, 
distilled from their complete 
publication history, are 
encoded in: {stable_str}

The research aspects most 
topically aligned with the 
target paper are encoded 
in: {dynamic_token}

{input_text}

Select the reference whose 
subject matter best matches 
this scholar's research domain. 
Output only [1] or [2].
\end{verbatim}
\end{tcolorbox}

\subsection{LaMP-2: Personalized Text Classification}
\begin{tcolorbox}[colback=gray!5, colframe=gray!40, title=Prompt Template, fonttitle=\bfseries]
\begin{verbatim}
A user's preferences learned 
from their past writings are 
encoded in: {stable_str}

The aspects of their history 
most relevant to the current 
task are encoded in: {dynamic_token}

{input_text}

Output only the answer.
\end{verbatim}
\end{tcolorbox}

\subsection{LaMP-3: Rating Prediction}
\begin{tcolorbox}[colback=gray!5, colframe=gray!40, title=Prompt Template, fonttitle=\bfseries]
\begin{verbatim}
This reviewer's personal rating 
calibration—how they translate 
expressed satisfaction 
into numerical scores—is encoded 
in: {stable_str}

The reviewer's past reviews whose 
sentiment most closely matches the 
current review 
are encoded in: {dynamic_token}

{input_text}

Based on this user's personal rating 
behavior, predict the score they 
assigned. 
Output only 1, 2, 3, 4, or 5.
\end{verbatim}
\end{tcolorbox}

\subsection{LaMP-4 / 5 / 7: Text Generation}
\begin{tcolorbox}[colback=gray!5, colframe=gray!40, title=Prompt Template, fonttitle=\bfseries]
\begin{verbatim}
This author's headline writing style 
and preferences, learned from their 
past articles, 
are encoded in: {stable_str}

The aspects of their past articles 
most relevant to the current article 
are encoded in: {dynamic_token}

{input_text}

Generate a headline in this author's 
personal writing style. 
Output only the headline.
\end{verbatim}
\end{tcolorbox}

\begin{figure}[t]
  \centering
  \includegraphics[width=\columnwidth]{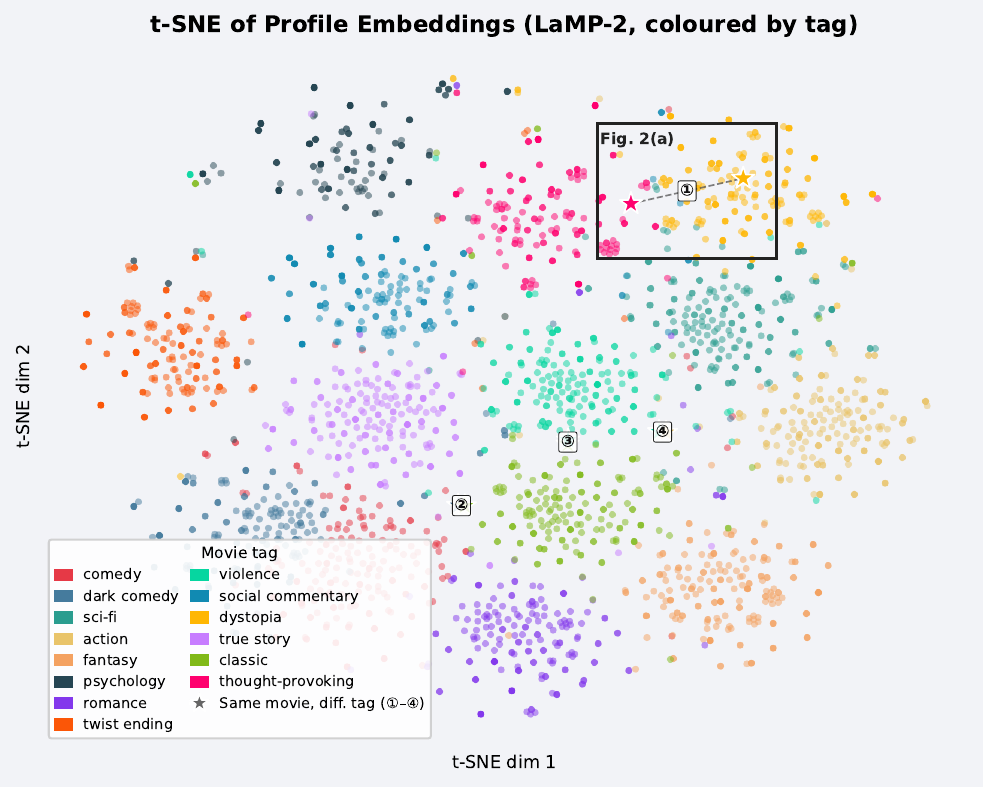}
  \caption{%
    t-SNE projection of 2{,}800 Contriever profile-item embeddings from
    \textsc{LaMP-2} (200 items $\times$ 14 movie tags), coloured by the
    user-assigned tag.
    Items with the same semantic content naturally cluster together,
    confirming that the learned PQ codebook partitions the embedding space
    along meaningful semantic axes.
    Stars \textcircled{1}--\textcircled{4} mark movies that appear in
    multiple users' profiles with \emph{different} tags
    (e.g., the same film is labelled \textit{dystopia} by one user and
    \textit{thought-provoking} by another);
    dashed lines connect each such pair.
    The rectangle is magnified in Figure~\ref{fig:codebook_zoom}.
  }
  \label{fig:codebook_tsne}
\end{figure}

\begin{figure}[t]
  \centering
  \includegraphics[width=\columnwidth]{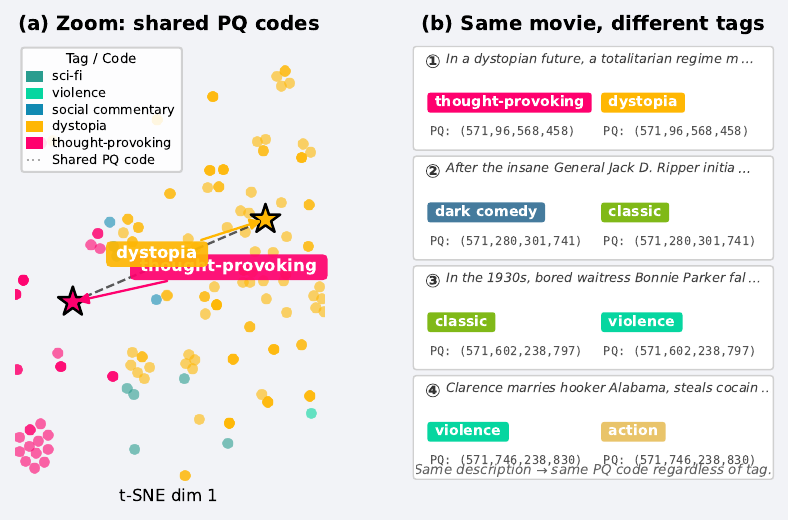}
  \caption{%
    \textbf{(a)} Zoom into the boxed region of Figure~\ref{fig:codebook_tsne}.
    Dotted lines connect items that share an \emph{identical} PQ code-tuple
    despite carrying different user-assigned tags, showing that the codebook
    groups items by content rather than by label.
    \textbf{(b)} Four concrete ``same movie, different tag'' pairs sampled
    from the training set.
    Each pair consists of the same movie description annotated by two
    different users with different tags.
    Because the description text is identical, the Contriever encoder
    produces the same embedding, and the PQ codebook assigns the same
    4-tuple code regardless of the user's label—demonstrating that the
    codebook captures \emph{content-level} semantics rather than
    user-specific label noise.
  }
  \label{fig:codebook_zoom}
\end{figure}

\section{Detailed Codebook Interpretability Analysis} \label{sec:inter}

Table~\ref{tab:pq_indices} provides detailed examples of PQ indices from the LaMP-4, demonstrating how the codebook decouples semantic dimensions. All samples share the same indices \textbf{[910, 204]} in subspaces 2 and 3, serving as a semantic anchor for Donald Trump-related content. Variations in subspaces 1 and 4 reveal fine-grained semantic distinctions in content type, perspective, sentiment, and stance.

  \begin{table*}[h]
  \centering
  \footnotesize
  \caption{PQ Indices Analysis for Trump Muslim Ban / Travel Ban Headlines (LaMP-4).
  All samples share indices \textbf{[40, 101]} in subspaces S2 and S3, forming a semantic
  anchor for Trump's Executive Order on immigration. Variations in S1 and S4 reveal
  fine-grained distinctions in coverage angle, source type, and perspective.}
  \label{tab:pq_indices}
  \begin{tabular}{p{2.0cm}p{4.5cm}p{3.8cm}cccc}
  \hline
  Sample & Text & Title & \multicolumn{4}{c}{PQ Indices} \\
   & & & S1 & S2 & S3 & S4 \\
  \hline
  \multicolumn{7}{l}{\textbf{Group 1: S1=20 (Legal/Political), S4=29 (Court Proceedings)}} \\
  308421-2815421 & And all because the Supreme Court could soon determine the legality of \textbf{Trump}'s executive
   order. & Travel Ban Challengers Lose Bid For Giuliani's `Muslim Ban' Memo & \textbf{20} & \textbf{40} &
  \textbf{101} & \textbf{29} \\
  308421-2815431 & The president's campaign promises are proving to be a key issue in the travel ban litigation. &
  \textbf{Trump}'s `Muslim Ban' Pledge Scrubbed From Website Just As Judges Ask About It & \textbf{20} & \textbf{40}
   & \textbf{101} & \textbf{29} \\
  308421-2815464 & A judge in Maryland agreed that anti-Muslim sentiment was behind the president's executive
  orders. & Federal Court In Hawaii Blocks \textbf{Donald Trump}'s New Travel Ban Nationwide & \textbf{20} &
  \textbf{40} & \textbf{101} & \textbf{29} \\
  308421-2815488 & The administration is trying to make the case that judicial review of its executive order
  imperils national security. & \textbf{Trump} Lawyers To Court Reviewing Muslim Travel Ban: Stay Out Of It &
  \textbf{20} & \textbf{40} & \textbf{101} & \textbf{29} \\
  \hline
  \multicolumn{7}{l}{\textbf{Group 2: S1=8 (Media/Civil Society), S4=65 (Personal Impact \& Press Critique)}} \\
  302244-745229 & They've been told to downplay the ``Muslim exclusion'' part of the Muslim exclusion order. & Wall
  Street Journal Editor Directs Reporters To Get Mealy-Mouthed Covering \textbf{Trump} & \textbf{8} & \textbf{40} &
  \textbf{101} & \textbf{65} \\
  304846-1424183 & A new report shows a dearth of Muslim voices in discussions about \textbf{Donald Trump}'s
  executive order. & Cable News Sure Could Talk To More Muslims About The Muslim Ban & \textbf{8} & \textbf{40} &
  \textbf{101} & \textbf{65} \\
  304846-1424184 & They were supposed to get married in Virginia and start a life together. Then \textbf{Donald
  Trump} signed an executive order. & How The Muslim Ban Is Ripping This Young Couple Apart & \textbf{8} &
  \textbf{40} & \textbf{101} & \textbf{65} \\

  \hline
  \multicolumn{7}{l}{\textbf{Group 3: S1=8 (Media/Civil Society), S4=29 (Constitutional Challenges)}} \\
  308421-2815426 & An appeals court said the president's executive order ``drips with religious intolerance.'' &
  Travel Ban Challengers Demand \textbf{Trump} Hand Over Giuliani's `Muslim Ban' Memo & \textbf{8} & \textbf{40} &
  \textbf{101} & \textbf{29} \\ 
  308421-2815490 & It's perhaps the most sweeping constitutional challenge to the president's travel ban yet. &
  Muslim Coalition Takes \textbf{Donald Trump} To Court Over `Muslim Exclusion Order' & \textbf{8} & \textbf{40} &
  \textbf{101} & \textbf{29} \\
  3011697-3528095 & A federal judge temporarily stopped elements of the executive order on Saturday. & Read The
  Court Order Halting Parts Of \textbf{Trump}'s Anti-Muslim Ban & \textbf{8} & \textbf{40} & \textbf{101} &
  \textbf{29} \\
  \hline
  \multicolumn{7}{l}{\textbf{Group 4: S1=20 (Legal/Political), S4=39 (Corporate Protest)}} \\
  3010214-3138252 & Googlers united against the president's decision to restrict immigration and travel from certain
   predominantly Muslim countries. & Google Workers Stage Large-Scale Walkout To Protest \textbf{Trump}'s Executive
  Order On Immigration & \textbf{20} & \textbf{40} & \textbf{101} & \textbf{39} \\
  \hline
  \multicolumn{7}{l}{\textbf{Group 5: S1=122 (Judicial Expert), S4=58 (Appellate Scrutiny)}} \\
  308421-2815432 & A key question an appeals court must answer is whether the president's anti-Muslim statements can
   be held against him. & Judges Hit \textbf{Donald Trump}'s Lawyer Hard About Legality Of Travel Ban & \textbf{122}
   & \textbf{40} & \textbf{101} & \textbf{58} \\
  \hline
  \end{tabular}
  \end{table*}

\textbf{Key Observations:}

  \begin{itemize}
      \item \textbf{Semantic Anchor (Subspaces 2 \& 3):} All samples share \textbf{[40, 101]}, capturing the core
  topic semantics (Trump's Muslim Ban / Travel Ban executive order content). This anchor spans 1,059 profile items
  with 99.2\% topic purity across the entire 3.6M-item training corpus.

      \item \textbf{Content Type (Subspace 1):}
      \begin{itemize}
          \item \textbf{20}: Legal/political proceedings (\textbf{Supreme Court}, \textbf{travel ban litigation},
  \textbf{Trump lawyers}, \textbf{judicial review}, \textbf{executive order})
          \item \textbf{8}: Media and civil society coverage (\textbf{Wall Street Journal}, \textbf{cable news},
  \textbf{Muslim voices}, \textbf{personal impact story}, \textbf{civil rights organizations})
          \item \textbf{122}: Expert appellate scrutiny (\textbf{appeals court judges}, \textbf{anti-Muslim
  statements}, \textbf{constitutional legality}, \textbf{lawyer cross-examination})
      \end{itemize}

      \item \textbf{Perspective/Framing (Subspace 4):}
      \begin{itemize}
          \item \textbf{29}: Court challenge framing (\textbf{challengers}, \textbf{filibuster}, \textbf{halting},
  \textbf{blocking}, \textbf{constitutional challenge})
          \item \textbf{65}: Personal impact and press accountability (\textbf{ripping couples apart},
  \textbf{dearth of Muslim voices}, \textbf{mealy-mouthed coverage}, \textbf{human cost})
          \item \textbf{39}: Protest/resistance framing (\textbf{Google workers walkout}, \textbf{large-scale
  protest}, \textbf{civil action})
          \item \textbf{58}: Expert legal proceedings framing (\textbf{judges hit lawyer hard}, \textbf{appellate
  questioning}, \textbf{legality hearing})
      \end{itemize}
  \end{itemize}

  This decomposition demonstrates that the PQ codebook learns meaningful semantic factorization. Notably, Groups 1
  and 3 share the same \textbf{S4=29} (court challenge framing) but differ in \textbf{S1} (20 vs.\ 8), separating
  government-side judicial proceedings from civil-society constitutional challenges---a nuanced distinction captured
   purely by the content-type subspace. This confirms that different subspaces capture hierarchical and orthogonal
  semantic dimensions: topic anchor (subspaces 2--3), content type/source (subspace 1), and perspective/framing
  (subspace 4).

\section{User Study}
  To understand what each model actually captures, we sample six representative
  users across all four tasks and compare predictions across the full baseline
  family, ZS, ICL, BM25, Dense, PPlug, PROPER, and the CURP ablations
  \textit{stable-only} and \textit{dynamic-only}against the full model
  (Table~\ref{tab:user_study}).

  \smallskip\noindent\textbf{What retrieval baselines capture through in-context
  copying.}
  ICL prepends the user's eight profile items verbatim as few-shot exemplars;
  BM25 and Contriever retrieve the most similar profile items and inject them as
  context.
  This gives retrieval baselines an immediate surface advantage: for the
  abbreviation-heavy Twitter user (LaMP-7, row~1), ICL, BM25, and Contriever all
  output \textit{``why didn't \textbf{u} tell me''} because the user's actual
  tweets (\textit{``I told u already''}, \textit{``why u gotta go2 work''}) appear
  verbatim in the prompt.
  Likewise in LaMP-4 (row~5), ICL (\textit{``8 Inappropriate\ldots''}), BM25
  (\textit{``8 Cringe-Worthy\ldots''}), and Contriever (\textit{``10 Wedding
  Comments\ldots''}) all spontaneously adopt the user's numbered-list format
  because numbered headlines appear directly in the retrieved context window.
  The mechanism is shallow: the model copies what it sees.
  When the query departs from the retrieved exemplars, as in the
  hallucination case (row~2) where no profile tweet is topically relevant,
  retrieval baselines produce a generic, coherent paraphrase rather than
  degenerating, because they never inject the profile into the generation
  pathway beyond the text prompt.

  \smallskip\noindent\textbf{Where learned representations outperform retrieval:
  conceptual genre framing.}
  The most discriminating case is LaMP-2, row~3, where an ambiguous film
  (a near-future vampire plague) can be legitimately tagged either
  \textit{sci-fi} or \textit{dystopia}.
  Every retrieval-based model, ICL, BM25, Contriever, and PPlug, predicts
  \textit{sci-fi}, because they retrieve the user's sci-fi movies from their
  profile and surface that label.
  Yet the user has tagged four films as \textit{dystopia} and only two as
  \textit{sci-fi}: the genre that dominates their \emph{conceptual} framing is
  not recoverable by retrieving the single most similar movie.
  Only PROPER, \textit{stable-only}, and CURP predict \textit{dystopia}
  correctly, because all three learn something about the user's overall preference
  distribution rather than nearest-neighbour matching.
  This is the clearest evidence that CURP's gains in LaMP-2 reflect genuine
  personalisation: it resolves lexically ambiguous films by internalising what
  genre the user reaches for, not by copying labels from retrieved items.

  \smallskip\noindent\textbf{When dynamic content overrides stable prior.}
  In LaMP-2 row~4, a user whose dominant tag is \textit{violence} ($\times$\,7
  films) must label a WWIII teen-survival story.
  The stable prior alone is insufficient: \textit{stable-only} predicts
  \textit{action}, the plausible surface reading, as do Direct, ICL, and BM25.
  The dynamic branch retrieves the user's violence-tagged war films and confirms
  \textit{violence}; CURP and PPlug both follow this content signal correctly.
  Notably, PROPER fails here (predicting \textit{comedy}), suggesting that
  PROPER's nearest-neighbour retrieval selects the wrong profile subset when the
  target film's description does not closely match any of the user's violence
  films in embedding space.

  \smallskip\noindent\textbf{A failure mode shared by all query-aware dynamic models.}
  Row~2 (LaMP-7) shows a systematic weakness of profile-embedding injection.
  When no profile tweet matches the paraphrase query, \textit{dynamic-only}
  degenerates into a hallucinated list of \texttt{@}handles
  (\textit{``@jessicajayy @michelle\_1986 @sarahjane1986\ldots''});
  PPlug exhibits the same failure (\textit{``@jessicajayy @michellebryant
  @sarahmccarthy\ldots''}).
  The retrieval baselines do not hallucinate here because they insert the profile
  as text: lacking relevant examples, the model simply ignores them.
  This failure is intrinsic to soft-prompt injection: without a relevant
  profile--query match, the injected embedding activates @mention patterns from
  the profile rather than suppressing output.
  \textit{Stable-only}, PROPER and CURP are the only personalised models to produce a
  coherent paraphrase on this example.

  \smallskip\noindent\textbf{Style and metrics decouple.}
  Row~5 (LaMP-4) is the clearest illustration.
  The gold reference headline contains no number
  (\textit{``Rudest Things Guests Say to the Bride and Groom''}),
  yet the user's entire profile follows a numbered format.
  Every personalised model, ICL (\textit{``8 Inappropriate\ldots''}),
  BM25 (\textit{``8 Cringe-Worthy\ldots''}), Contriever, PROPER, Stable,
  Dynamic, and CURP predicts a numbered headline more consistent with the
  user's observable writing habit than the reference itself.
  All are penalised by ROUGE.
  This confirms that automatic metric gains systematically undercount style
  improvements: a model that writes the way the user always writes is penalised
  whenever one specific reference deviates from that habit.

Overall, by case study, we demonstrate our CURP model can achieve a more personalized performance beyond the automatic metric and keep a balance on self-inner stable representation and query-aware dynamic representation.

\begin{table*}[t]
  \centering
  \footnotesize
  \caption{%
    Qualitative comparison across models and cases. 
    \textbf{Bold} indicates critical differentiating tokens.
    $\checkmark$ = style-consistent or correct; $\times$ = inconsistent or incorrect.
  }
  \label{tab:qualitative_transposed_simple}
  
  \setlength{\tabcolsep}{2.5pt} 
  \renewcommand{\arraystretch}{1.2}
  
  \begin{tabular}{
    l 
    c c c c c c
  }
  \toprule
  \textbf{Model} & 
  \textbf{R1} & 
  \textbf{R2} & 
  \textbf{R3} & 
  \textbf{R4} & 
  \textbf{R5} & 
  \textbf{R6} \\
  
  \multicolumn{1}{l}{\scriptsize \textit{(Case Context)}} & 
  \scriptsize LaMP-7 & 
  \scriptsize LaMP-7 & 
  \scriptsize LaMP-2 & 
  \scriptsize LaMP-2 & 
  \scriptsize LaMP-4 & 
  \scriptsize LaMP-3 \\
  
  \multicolumn{1}{l}{\scriptsize \textit{(Trait)}} & 
  \scriptsize Abbrev. & 
  \scriptsize Sparse & 
  \scriptsize Genre & 
  \scriptsize Content & 
  \scriptsize Format & 
  \scriptsize Rating \\
  \midrule

  \textbf{Direct} &
  \ldots \textbf{you} $\times$ &
  \textit{coh.} $\checkmark$ &
  fantasy $\times$ &
  action $\times$ &
  \textit{Wedding Woes} $\times$ &
  \textbf{5} $\times$ \\

  \textbf{ICL} &
  \ldots \textbf{u} $\checkmark$ &
  \textit{coh.} $\checkmark$ &
  \textbf{sci-fi} $\times$ &
  action $\times$ &
  \textbf{8} Inapp. $\checkmark$ &
  \textbf{5} $\times$ \\

  \textbf{BM25} &
  \ldots \textbf{u} $\checkmark$ &
  \textit{coh.} $\checkmark$ &
  \textbf{sci-fi} $\times$ &
  action $\times$ &
  \textbf{8} Cringe $\checkmark$ &
  \textbf{5} $\times$ \\

  \textbf{Cont.} &
  \ldots \textbf{u} $\checkmark$ &
  \textit{coh.} $\checkmark$ &
  \textbf{sci-fi} $\times$ &
  sci-fi $\times$ &
  \textbf{10} Wedding $\checkmark$ &
  4 $\checkmark$ \\

  \textbf{PPlug} &
  \ldots \textbf{u} $\checkmark$ &
  @jessica\ldots $\times$ &
  \textbf{sci-fi} $\times$ &
  violence $\checkmark$ &
  \textit{Etiquette} $\times$ &
  4 $\checkmark$ \\

  \textbf{PROPER} &
  \ldots \textbf{you} $\times$ &
  \textit{coh.} $\checkmark$ &
  dystopia $\checkmark$ &
  \textbf{comedy} $\times$ &
  \textbf{10} Wedding $\checkmark$ &
  \textbf{5} $\times$ \\

  \textbf{Stable} &
  \ldots \textbf{you} $\times$ &
  \textit{coh.} $\checkmark$ &
  dystopia $\checkmark$ &
  \textbf{action} $\times$ &
  \textbf{10} Wedding $\checkmark$ &
  4 $\checkmark$ \\

  \textbf{Dynamic} &
  \ldots \textbf{you} $\times$ &
  @jessica\ldots $\times$ &
  \textbf{sci-fi} $\times$ &
  violence $\checkmark$ &
  \textbf{10} Wedding $\checkmark$ &
  4 $\checkmark$ \\

  \textbf{CURP} &
  \ldots \textbf{u} $\checkmark$ &
  \textit{coh.} $\checkmark$ &
  dystopia $\checkmark$ &
  violence $\checkmark$ &
  \textbf{10} Wedding $\checkmark$ &
  4 $\checkmark$ \\

  \midrule
  \textbf{Ref.} &
  \ldots \textbf{u} tell me &
  \textit{Good morning\ldots} &
  dystopia &
  violence &
  \textit{Rudest Things\ldots} &
  4 \\

  \bottomrule
  \end{tabular}

  \vspace{0.5em}
  \parbox{\linewidth}{\scriptsize
    \textbf{R1 (Abbrev.):} CURP/ICL/BM25 capture ``u'' (style); Direct/PROPER/Stable/Dyn use ``you''. \quad
    \textbf{R2 (Sparse):} Retrieval models remain coherent; Embedding models (PPlug/Dyn/CURP) hallucinate handles. \quad
    \textbf{R3 (Genre):} CURP/PROPER/Stable infer ``dystopia'' (global prior); others match local ``sci-fi'' items. \quad
    \textbf{R4 (Content):} CURP/PPlug/Dyn retrieve ``violence''; PROPER/Stable miss; others guess ``action/comedy''. \quad
    \textbf{R5 (Format):} All personalized models adopt numbered format (style-consistent); Reference is unnumbered. \quad
    \textbf{R6 (Rating):} CURP/Cont./PPlug/Stable/Dyn calibrate to 4; Direct/ICL/BM25/PROPER overshoot to 5.
  }
  \label{tab:user_study}
\end{table*}

\section{Combine with PEFT}

As shown in Table~\ref{tab:lora_ablation}, integrating LoRA-based fine-tuning generally enhances the performance of CURP on structured and semantic tasks. Specifically, we observe consistent improvements on LaMP-2 (Classification), LaMP-3 (Rating Prediction), and LaMP-4 (Headline Generation), where LoRA helps the model better align with user-specific preferences and domain-specific patterns, yielding lower error rates (MSE/RMSE) and higher generation quality (ROUGE scores).
However, an interesting divergence occurs on LaMP-7 (Tweet Generation). Here, the vanilla CURP outperforms the LoRA-fine-tuned variant. We attribute this to the unique nature of tweet data, which is characterized by highly diverse, complex, and informal linguistic styles (e.g., slang, abbreviations, and erratic syntax). While LoRA effectively captures stable, long-term user profiles, full parameter freezing with lightweight adaptation may inadvertently smooth out the high-frequency, stochastic variations inherent in casual social media text. In such scenarios, the rigid structure imposed by LoRA fine-tuning might constrain the model's ability to reproduce the raw, unpolished authenticity required for realistic tweet generation, whereas the retrieval-augmented approach of CURP alone preserves these fine-grained stylistic nuances more effectively.

\section{Licenses}
We use publicly available datasets and models that are already licensed. Our code will be released if accepted with MIT License.

\end{document}